# A Non-Parametric Subspace Analysis Approach with Application to Anomaly Detection Ensembles

Marcelo Bacher[1], Irad Ben-Gal[1], Erez Shmueli[1*]

[1] Department of Industrial Engineering, Tel-Aviv University, Israel

[*] Corresponding author. E-Mail: shmueli@tau.ac.il, Address: Tel-Aviv University, Department of Industrial Engineering, Ramat Aviv, Tel-Aviv 69978, Israel.

**Abstract**: Identifying anomalies in multi-dimensional datasets is an important task in many real-world applications. A special case arises when anomalies are occluded in a small set of attributes, typically referred to as a *subspace*, and not necessarily over the entire data space. In this paper, we propose a new subspace analysis approach named Agglomerative Attribute Grouping (AAG) that aims to address this challenge by searching for subspaces that are comprised of highly correlative attributes. Such correlations among attributes represent a systematic interaction among the attributes that can better reflect the behavior of normal observations and hence can be used to improve the identification of two particularly interesting types of abnormal data samples: anomalies that are occluded in relatively small subsets of the attributes and anomalies that represent a new data class. AAG relies on a novel multi-attribute measure, which is derived from information theory measures of partitions, for evaluating the "information distance" between groups of data attributes. To determine the set of subspaces to use, AAG applies a variation of the well-known agglomerative clustering algorithm with the proposed multi-attribute measure as the underlying distance function. Finally, the set of subspaces is used in an ensemble for anomaly detection. Extensive evaluation demonstrates that, in the vast majority of cases, the proposed AAG method (i) outperforms classical and state-of-the-art subspace analysis methods when used in anomaly detection ensembles, and (ii) generates fewer subspaces with a fewer number of attributes each (on average), thus resulting in a faster training time for the anomaly detection ensemble. Furthermore, in contrast to existing methods, the proposed AAG method does not require any tuning of parameters.

## 1 Introduction

Anomaly detection refers to the problem of finding patterns in data that do not conform to an expected norm behavior. These non-conforming data points or patterns are often referred to as anomalies, outliers, discordant observations, exceptions, aberrations, surprises, peculiarities or contaminants, depending on the application domain (Chandola et al., 2007). Algorithms for detecting anomalies are extensively used in a wide variety of application domains, such as machinery monitoring (Ben-Gal et al., 2003; Ge and Song, 2013; Kenett and Zacks, 2014; and Bacher et al. 2017), sensor networks

(Bajovic et al., 2011), intrusion detection in data networks (Jyothsna et al., 2011), health care (Tarassenko et al., 2005), and social networks (Aggarwal et al., 2012). A major reason for their widespread use is the fact that, in many cases, anomalies can be translated directly to actionable recommendations based on either "good" or "bad" deviations from the norm (Chandola et al., 2007).

In a typical anomaly detection setting, only normal or expected observations are available, and consequently, some assumptions regarding the distribution of anomalies must be made to discriminate normal from anomalous observations (Steinwart et al., 2006). Traditional approaches for anomaly detection (see, e.g., Ben-Gal et al., 2010; and Pimentel et al., 2014) often assume that anomalies occur sporadically and are well separated from the normal data observations or that anomalies are uniformly distributed around the normal observations. However, in complex environments, such assumptions may not hold. For instance, consider the case of a complex system and a diagnosis module that continuously monitors the functionality of the system by analyzing multi-attribute (we use the terms attribute, variable and feature interchangeably) data generated from a set of sensors. If only one of the system's modules breaks down, or alternatively, if only a few of the monitoring sensors fail to function normally, only some of the data attributes will be affected. From a data analysis perspective, these malfunctions can be seen as an addition of noise to some subset of the attributes. Consequently, anomalies in the system's generated data might only be noticeable or visible in some projections of the data into a lower-dimensional space, typically called a subspace, and not necessarily in the entire data space, as often assumed by classical approaches. As another motivating example, consider a case where anomalies represent a new, previously unknown, class of data observations, commonly called novelties (Chandola et al., 2007). Similar to the malfunctions example above, deviations from the original data observations might only be visible along a subset of attributes. However, these attributes will often be correlated in some sense and therefore cannot be treated as additive noise.

Based on these concepts, ensembles were proposed as a relatively new paradigm for anomaly detection (Aggarwal and Yu, 2001). Ensembles for anomaly detection typically follow three general steps (Lazarevic and Kumar, 2005). First, a set of subspaces is generated (often by randomly selecting subsets of attributes). This step is commonly referred to as *subspace analysis*. Then, classical anomaly detection algorithms are applied on each subspace to compute local anomaly scores. Finally, these local scores are aggregated to derive a global anomaly score (e.g., using majority voting). Here, we focus on the subspace analysis stage, which aims to find a representative set of subspaces among a very large number of possible subspace combinations such that anomalies can be identified effectively and efficiently.

Several methods for subspace analysis have been proposed in the literature. These methods can be classified into three broad approaches. The most basic one is based on a random selection of attributes (e.g., Lazarevic and Kumar, 2005). Other methods search for subspaces by giving anomality grades to data samples, thus coupling the search for meaningful subspaces with the anomaly detection algorithm (see, e.g., Müller et al., 2010; and Ha et al., 2015). Recent methods search for subspaces comprising of highly correlative attributes (e.g., Nguyen et al., 2014). These methods rely on the assumption that, in such subspaces, the correlations among attributes represent a systematic interaction among the attributes that can better reflect the behavior of normal observations and hence can be used to better identify those deviating abnormal cases. However, all of the above methods suffer from one or more of the following limitations: (i) Relevant attributes might not be included in the generated set of subspaces. This might impact the effectiveness of the ensemble since anomalies might occur anywhere in the data space; (ii) The set of generated subspaces might contain thousands of subspaces, which may make the training and operation phases of the ensemble computationally prohibitive; (iii) These approaches often require, prior to their execution, to set the values of parameters such as the number of subspaces, the maximal size of each subspace or the number of clusters, that are typically hard to predefine or tune at such a stage.

To address the challenges mentioned above, we propose the Agglomerative Attribute Grouping method (AAG) for subspace analysis. Motivated by previous works, AAG searches for subspaces that are comprised of highly correlative attributes. As a general measure for attribute association, AAG applies a measure derived from information-theory measures of partitions (see, Simovici, 2007; and Kagan and Ben-Gal, 2014). In particular, AAG uses the Rokhlin distance (Rokhlin, 1967) to evaluate the smallest distance between subspaces in the case of two attributes and a multi-attribute measure, which is proposed here, for cases where more than two attributes are involved. Then, AAG applies a variation of the well-known agglomerative clustering algorithm where subspaces are greedily searched by minimizing the multi-attribute measure. We also propose a pruning mechanism that aims at improving the convergence time of the proposed algorithm, while limiting the size of the subspaces.

Several important characteristics differentiate AAG from existing state-of-the-art approaches. First, due to the used agglomerative scheme in the subspace search, none of the data attributes are discarded, and attributes are combined in an effective manner to generate the set of subspaces. Second, the set of subspaces that AAG generates is relatively "compact" in comparison to existing methods for two main reasons: the use of the agglomerative approach results in a relatively small number of subspaces, and the pruning mechanism results in a relatively small number of attributes in each subspace. Finally, as a result of combining the agglomerative approach with the minimization of the suggested measure, AAG does not require any tuning of parameters.

To evaluate the proposed AAG method, we conducted extensive experiments on 25 publicly available datasets while using eight different classical and state-of-the-art subspace analysis methods as benchmarks. The evaluation results show that an AAG-based ensemble for anomaly detection: (i) outperforms the benchmark methods in cases where anomalies occur in relatively small subsets of the available attributes, as well as in cases where anomalies represent a new class (i.e., novelties); and (ii) generates fewer subspaces with a smaller (on average) number of attributes in comparison to the benchmark approaches, thus resulting in a faster training time for the anomaly detection ensemble.

It is important to note that, while subspace analysis for anomaly detection seems to be similar to attribute selection for supervised classification (Guyon et al., 2006), as well as to some ensemble-based classification methods (e.g., Random Forest in Breiman, 2001), they differ greatly. The main difference between the two approaches stems from the type of data available in the training phase of the classification task vs. the available data for the anomaly detection task. While in the supervised classification task, information about each of the classes is usually available, in anomaly detection tasks, information about abnormal data samples is often missing and only information about the normal observations is provided. Moreover, while the goal in the case of attribute selection is to discard redundant attributes to improve accuracy and run-time of the classifier, it is usually impossible to discard attributes at the training stage of anomaly detection tasks since they might be found to be extremely informative in the operational stage for detecting anomalies.

The contribution of this paper is twofold. First, it introduces a new multi-attribute information theoretic measure, which can be seen as an extension to the Rokhlin metric. The proposed measure enables to compute the expected information gain of potential subspaces, with the aim of identifying unexpected observations. The new measure has several appealing properties: (i) unlike many other measures, such as Pearson correlation, it can be computed over a set of more than two variables; ii) unlike other measures such as Pearson correlation that can handle numerical attributes only, it can handle numerical as well as categorical variables; and (iii) it enables to expose high-order nonlinear dependencies among attributes, while simpler correlation measures often reveal linear dependencies between variables. The latter property is further addressed in Appendix A. To the best of our knowledge, this paper is the first one to use the Rokhlin distance and its multi-attribute extension, in the context of subspace analysis.

Second, this paper introduces the Agglomerative Attribute Grouping method (AAG), which is a novel algorithm for subspace analysis. The proposed AAG is unique in the sense that: (i) it is non-parametric, (ii) it outperforms other methods when used in anomaly and novelty detection ensembles; and (iii) it runs faster than other methods.

This work extends two earlier conference papers (Bacher et al., 2016, 2017) by: (i) expanding the selection mechanism of AAG to support a stability index for the selected subspaces; (ii) outlining properties of the proposed measure and proving them; (iii) providing an extensive evaluation of the proposed approach which now includes additional datasets and benchmarks; and (iv) elaborating the statistical analyses of the obtained results.

The rest of the paper is organized as follows. Section 2 discusses previous works related to subspace analysis. Section 3 proposes a novel measure, based on concepts of information-theory over sets of partitions, that enables to evaluate the smallest "distance" among subspaces of different attributes. Section 4 describes the proposed AAG approach. Section 5, presents an experimental evaluation of AAG and the obtained results. Finally, Section 6 summarizes this paper and discusses some future research directions.

## 2 Related Work

In this section, we review classical approaches for anomaly detection, while focusing on ensemble-based approaches. We mainly consider subspace analysis methods for anomaly detection; however, we also review some closely related works, in which subspace analysis is used for clustering purposes.

### 2.1 Classical Anomaly Detection Approaches

In Machine Learning applications, anomaly detection methods aim at detecting data observations that considerably deviate from normal data observations (Aggarwal, 2015). Anomaly detection techniques can be broadly classified into four categories, based on the type of data available during the training stage: (i) Supervised anomaly detection techniques that assume the availability of labeled instances for both normal and abnormal classes. These techniques are similar to those used for classification of imbalanced datasets; (ii) One-class anomaly detection techniques for which the training data consists only of instances associated with the normal class. These methods do not require labeled instances for the anomaly class, therefore they are more widely applicable than the supervised techniques; (iii) Semi-supervised anomaly detection techniques that assume a relatively small number of labeled instances and a large number of unlabeled instances. One notable setting of semi-supervised techniques is PU learning, in which all labeled instances are associated with the normal class (this setting is also very similar to that of the one-class anomaly detection, only that its training data contains also unlabeled instances); and iv) Techniques that operate in a full unsupervised mode and do not require any labelled training data. These techniques often make the implicit assumption that normal instances are far more frequent than anomalies in the test data. If this assumption is not true, then such techniques often result in high false alarm rate. In this work we focus on one-class anomaly detection techniques only. For well-documented surveys on anomaly detection techniques, the reader is referred to the literature (e.g., Markou and Singh, 2003; Chandola et al., 2007; Ben-Gal, 2010; and Pimentel et al., 2014).

While classical anomaly detection techniques are widely used in real-world applications, they share a major limitation: the underlying assumption that abnormal observations are sporadic and isolated with respect to normal data samples. That is, abnormal observations are usually seen as the result of additive random noise in the full data space (Chandola et al., 2007). Under this assumption, anomalies can often be identified by building a single model that describes normal data along all of its dimensions. However, in complex environments, such assumptions may not hold. Specifically, abnormal data samples might be occluded in some subspaces that represent combinations of attributes[1], and may only be noticed in lower-dimensional projections, or subspaces. Consequently, classical approaches that examine the entire data space, are less fitted for identifying abnormal data samples in these settings.

## 2.2 Anomaly Detection Ensembles

One of the first approaches that aimed at identifying relevant subspaces was presented in (Aggarwal and Yu, 2001). Several works followed this direction by proposing enhanced methods where data observations were ranked based on the number of subspace projections they were visible in to define their "anomality grade" (see, e.g., Kriegel et al., 2009; Müller et al., 2010; and Müller et al., 2011). Note, however, that in complex settings, such an approach may not hold, since abnormal data samples might be occluded in very specific and potentially low dimensional subspaces only. One limitation of these approaches is the assumption that anomalous observations are mixed together with normal data samples, and therefore, the resulting set of subspaces depends on the anomality grade of each observation. Another limitation is that the anomaly detection algorithm is tightly coupled with the search strategy for subspaces (Keller et al., 2012).

A different subspace search mechanism for anomaly detection was presented by Lazarevic and Kumar (Lazarevic and Kumar, 2005) and named Feature Bagging (FB). In FB, attributes were first selected, in a random manner, to generate subspaces of different sizes. Then, a classical anomaly detection algorithm (see, e.g., Aggarwal, 2015) was trained in each subspace and then aggregated to derive a global novelty score per observation. The naive random selection of subspaces has the drawback that some attributes might not be selected at all. Consequently, high detection performance cannot be guaranteed. However, the FB approach proposed an inspiring framework that divided the anomaly detection challenge into three main stages: subspace analysis, anomality score computation and score aggregation. Thus, the task of finding "good" subspaces can be isolated from the anomaly detection algorithm, as well as from the strategy for aggregating scores that are used at later stages.

---

[1] Note that such anomalies share some similarity with contextual anomalies (Chandola et al., 2007), where the subspace represent the set of contextual attributes.

Further works that focused on the search for subspaces were presented in (Keller et al., 2012; Nguyen et al., 2013; and Nguyen et al., 2014). In Keller et al., (2012), High Contrast Subspaces (HiCS) was presented as a method to search for combinations of attributes based on the A-Priori Algorithm (Agrawal and Srikant, 1994) and on a random permutation of potentially useful attributes. Attributes are selected to form a subspace by quantifying the differences between their marginal and conditional distributions, thus looking for subspaces with high information gain. Due to the applied search strategy, HiCS can potentially generate thousands of subspaces, which makes the training phase of the ensemble for anomaly detection computationally prohibitive. To overcome this inherent drawback, it is necessary to reduce the number of generated subspaces. However, this is not a trivial task since arbitrarily selecting a smaller subset of these subspaces may result in poorer performance of the ensemble.

Nguyen et al. presented two approaches, Cumulative Mutual Information (CMI) (Nguyen et al., 2013) and 4S (Nguyen et al., 2014). In CMI, subspaces are searched using the conditional mutual information as the difference between the joint distribution of several random variables and the product of their marginals. CMI starts with two-dimensional subspaces, where in each step, it collects $M$ subspaces to generate new candidates in a level-wise manner. To effectively compute the conditional mutual information, the approach makes use of the $k$-means (MacQueen, 1967) clustering algorithm, where the number of clusters is previously defined.

Similar to CMI, the 4S approach combines subspaces by considering the maximal Total Correlation (Watanabe, 1960) among attributes computed by means of their cumulative distributions. The underlying algorithm starts with two attributes and then greedily searches for candidates that yield a higher correlation within the subspace. The Total Correlation is computed by using pairwise candidates, and then the values are sorted in descending order. Following this stage, the 4S method selects a predefined maximal number of attributes to be combined into a subspace. As no prior information regarding novelties usually exists, this selection strategy might lead the ensemble to performance deterioration because attributes might be discarded. Furthermore, selecting the maximal number of attributes in advance is mandatory since the Total Correlation is a monotonic non-decreasing function, and it therefore requires the size of the subspaces to be bounded. Unfortunately, selecting the right value for this particular parameter is far from trivial or robust since it has a significant impact on the resulting set of subspaces.

## 2.3  Subspace Analysis for Clustering

In addition to anomaly detection, several subspace analysis approaches have been proposed for data clustering. In particular, subspace clustering is an extension of conventional clustering that seeks to find clusters in different subspaces within a dataset (Parsons et al., 2004). The clusters are usually described by a group of attributes that contribute the most to the compactness of the data observations within the subspace. In (Deng et al., 2016), subspace clustering approaches are divided into two different groups: hard subspace clustering (HSC) and soft subspace clustering (SSC) algorithms.

HSC methods aim to find the exact set of attributes in each cluster. Aggarwal et al. presented CLIQUE in (Agrawal et al., 1998) as one of the first HSC methods. In (Cheng et al., 1999), based on concepts from (Agrawal et al., 1998), ENCLUS was presented as a method that searches for subspaces with computed low Shannon entropy (see, e.g., Cover and Thomas, 2006), and numerous works on subspace analysis for anomaly detection have been benchmarked (see, e.g., Keller et al., 2012; Nguyen et al., 2013; and Nguyen et al., 2014). As Shannon entropy measures the uncertainty level of the joint distribution over the subspace, searching for subspaces whose entropy results are lower than a threshold implies that the observations within the subspace follow a distribution that is different from the uniform, i.e., they represent a certain level of correlation among the attributes. However, the A-priori search strategy generates several hundreds of subspaces in a similar fashion to HiCS with the same time-consuming consequences for training and testing. A detailed review of HSC algorithms can be found in, e.g., (Parsons et al., 2004).

On the other hand, SSC algorithms perform the subspace clustering by assessing a weighting factor for each attribute in proportion to the contribution to the formation of a particular cluster. In general, SSC approaches compute weighting factors of the attributes using a $k$-means-based clustering algorithm (see, e.g., Gan et al., 2006; Jing et al., 2007; and Gan and Ng, 2015). In (Jing et al., 2007), for example, the authors introduced the Entropy Weighted $k$-means algorithm (EWKM), extending the classical $k$-means algorithm. EWKM constructs clusters by adopting the rule of maximizing entropy of the weighting distribution while adjusting these factors to the data attributes during the $k$-means optimization. More recently, Gan et al., introduced Automatic Feature Grouping k-means (AFG-k-means) (Gan and Ng, 2015) that proposed a new component into the objective function to automatically group attributes in addition to weighting them as a function of their clustering importance.

One of the major drawbacks of subspace clustering algorithms is the challenge of tuning specific parameters for each algorithm, as well as the identification of the correct number of clusters that the $k$-means algorithm requires prior to its execution.

## 3 Information Theory Measures for Partitions

This section discusses how to use information-theoretic measures over partitions of a generic dataset to compute the distances among various subsets of attributes. In particular, we review the Rokhlin distance (Rokhlin, 1967) and the symmetric difference (see, e.g., Kuratowski, 2014) and their application for attributes' association following a partitioning of a dataset. We then suggest an extension of the Rokhlin distance for a multi-attribute measure for any number of attributes. To that end and to maintain a self-contained text, we start this section by providing a brief review of concepts of partitions and their implementation to information theory while presenting the notation that is used throughout the paper.

### 3.1 Preliminaries

In this subsection, we follow (Kagan and Ben-Gal, 2014) and present some definitions of information-theoretic measures between partitions of a finite dataset. Let $D$ be a finite sample space composed of

$N$ observations and $p$ attributes: $\{A_1, A_2, ..., A_p\}$, and let $\chi$ be a set of partitions of the sample space $D$ as defined next. Each partition $\alpha_i$ is defined by the values of the corresponding attribute $A_i$, $\alpha_i = \{\alpha_{i1}, \alpha_{i2}, ..., \alpha_{iK}\}$, $K \leq N$, $\alpha_{ij} \cap \alpha_{im} = \emptyset$, $\forall j, m = 1, 2, ..., K$, $j \neq m$. For an attribute $A_i \in \{A_1, A_2, ..., A_p\}$, the elements $\alpha_{ij}$ of the corresponding partition $\alpha_i = \{\alpha_{i1}, \alpha_{i2}, ..., \alpha_{iK}\}$ are the sets of indices of unique values of the attribute $A_i$. For example, let the attribute $A_i$ contain the observations $\{a_{i1}, a_{i2}, ..., a_{iN}\}$, such that $a_{i1} = a_{i2} = a_{i3}$ and $a_{ij} \neq a_{ik}$ $\forall j, k = 3, 4, ..., N$, $j \neq k$. Then, the partition of $D$ generated by the attribute $A_i$ is $\{\{a_{i1}, a_{i2}, a_{i3}\}, \{a_{i4}\}, ..., \{a_{iN}\}\}$, which in terms of indices is represented by $\alpha_i = \{\alpha_{i1} = \{1,2,3\}, \alpha_{i2} = \{4\}, ..., \alpha_{iN-2} = \{N\}\}$. Note that, by definition, the union of the partition elements is the set of all indices, i.e., $\bigcup_{j=1}^{K} \alpha_{ij} = \{1, 2 ... N\}$, $\forall i$.

To define the entropy and the informational measures between partitions, rather than the conventional approach that defines them between random variables, it is necessary to specify a probability distribution associated with a partition. For finite sets, the empirical probability distribution induced by a partition $\alpha_i \in \chi$ is defined as follows (Simovici, 2007, Kagan and Ben-Gal, 2014):

$$p_{\alpha_i} = \left(\frac{|\alpha_{i1}|}{N}, \frac{|\alpha_{i2}|}{N}, ..., \frac{|\alpha_{iK}|}{N}\right),$$

where $|\cdot|$ represents the cardinality of the set. Note that by definition, $\sum_{j=1}^{K}(|\alpha_{ij}|/N) = 1$. Thus, the partition $\alpha_i$ induces a random variable $X_i = \{\alpha_{i1}, \alpha_{i2}, ..., \alpha_{iK}\}$, with the probabilities $p_{\alpha_i}(\alpha_{ij})$ defined over the partition elements $\alpha_{ij}$, $j = 1, 2, ..., K$. The Shannon entropy of the random variable $X_i$ of partition $\alpha_i$ is then defined as $H(X_i) = -\sum_{\alpha_{ij} \in \alpha_i} p_{\alpha_i}(\alpha_{ij}) \log_2[p_{\alpha_i}(\alpha_{ij})]$, where by the usual convention $0 \log_2 0 = 0$. Notice that the probabilities used in computing the entropy are obtained from the relative frequency of unique occurrences of the values of the attribute $A_i$. Therefore, one can compute the Shannon entropy of the partition $\alpha_i$ simply by $H(A_i)$.

Let $\alpha_i$ and $\beta_j$ be two partitions generated by the attributes $A_i$ and $A_j$ respectively, where $\alpha_i = \{\alpha_{i1}, \alpha_{i2}, ...\}$ and $\beta_j = \{\alpha_{j1}, \alpha_{j2}, ...\}$. Then, the conditional entropy of the partition $\alpha_i$ with respect to the partition $\beta_j$ is defined as follows:

$$H(\alpha_i|\beta_j) = -\sum_{\alpha_{jk} \in \beta_j} \sum_{\alpha_{im} \in \alpha_i} p(\alpha_{im}, \alpha_{jk}) \log_2[p(\alpha_{im}|\alpha_{jk})],$$

where $p(\alpha_{im}, \alpha_{jk}) = p(\alpha_{im} \cap \alpha_{jk})$ and $p(\alpha_{im}|\alpha_{jk}) = p(\alpha_{im} \cap \alpha_{jk})/p(\alpha_{jk})$. The Rokhlin distance between two partitions $\alpha_i$ and $\beta_j$ is then defined as the sum of conditional entropies between these partitions, that is (Rokhlin, 1967),

$$d_R(\alpha_i, \beta_j) = H(\alpha_i|\beta_j) + H(\beta_j|\alpha_i) \qquad (1)$$

For detailed consideration of the metric properties of this distance, see, (Sinaï, 1977). Recall that the partitions $\alpha_i$ and $\beta_j$ are associated with attributes $A_i$ and $A_j$, respectively. The rationale presented before allows us to compute the Shannon entropy of a partition $\alpha_i$ by computing the relative frequency of the symbols of the attribute $A_i$. Therefore, the conditional entropy $H(A_i|A_j)$ is also specified as the conditional entropy of the partition $\alpha_i$ with respect to the partition $\beta_j$. That is, $H(A_i|A_j) = H(\alpha_i|\beta_j)$. It follows, that the Rokhlin distance is equivalently computed as,

$$d_R(A_i, A_j) = H(A_i|A_j) + H(A_j|A_i)$$
$$= H(\alpha_i|\beta_j) + H(\beta_j|\alpha_i) = d_R(\alpha_i, \beta_j) \qquad (2)$$

Note that the Rokhlin distance is directly related to Shannon's mutual information as a measure of entropy reduction. Recall that $I(A_i; A_j) = H(A_i) - H(A_i|A_j)$ and $H(A_i) = H(A_i; A_j) - H(A_j|A_i)$ (Cover and Thomas, 2006). Thus, $d_R(A_i, A_j) = H(A_i, A_j) - I(A_i; A_j)$. Accordingly, the Rokhlin distance can be interpreted as a measure of mutual dependence between two attributes. A small Rokhlin distance reflects a small conditional entropy value and a high mutual information value between the dependent attributes. A direct implementation of the Rokhlin distance as a formal informational metric between partitions has practical implications. For example, it was used in (Kagan and Ben-Gal, 2013) for constructing a search algorithm and in (Kagan and Ben-Gal, 2014) for creating various testing trees.

As an illustrative numeric example, consider a dataset $D$ with $p = 2$ attributes (rows) and $N = 10$ observations (columns). Table 1 shows the transposed dataset $D$, where attribute $A_1$ takes on binary values and attribute $A_2$ takes on the colors Blue (B), Red (R) or Green (G).

Table 1: An example of a dataset $D$ with $N = 10$ and $p = 2$.

| $A_1$ | 0 | 1 | 0 | 1 | 0 | 0 | 1 | 0 | 0 | 1 |
|---|---|---|---|---|---|---|---|---|---|---|
| $A_2$ | R | G | R | G | G | R | B | B | R | G |

The partition of $A_1$ results in $\alpha_1 = \{\{1,3,5,6,8,9\}, \{2,4,7,10\}\}$ with corresponding probability values $p_1 = (0.6, 0.4)$. Similarly, the partition of $A_2$ results in $\alpha_2 = \{\{7,8\}, \{1,3,6,9\}, \{2,4,5,10\}\}$, with corresponding probability values $p_2 = (0.2, 0.4, 0.4)$. The entropy for the partition $\alpha_1$ is computed as $H(\alpha_1) \cong 0.97$, and the entropy for partition $\alpha_2$ is computed as $H(\alpha_2) \cong 1.52$. The conditional entropy $H(\alpha_1|\alpha_2) = H(A_1|A_2) = 0.2 \times H(A_1|A_2 = B) + 0.4 \times H(A_1|A_2 = R) + 0.4 \times H(A_1|A_2 = G) \cong 0.525$, and $H(A_2|A_1) \cong 1.076$. Finally, the Rokhlin distance is computed as $d_R(A_1, A_2) \cong 1.60$.

## 3.2 Multi-Attribute Measure

Equation (2) measures the informational distance between two attributes. Subsequently, we now extend this concept to derive a similar notion of distance between a set of multiple attributes. The new *multi-attribute measure* among attributes is induced by sets of partitions and is denoted by $d_{MA}$.

The symmetric difference between two partitions, which is also known as the *disjunctive union*, is defined as follows: $\alpha_i \Delta \beta_j = (\alpha_i \backslash \beta_j) \cup (\beta_j \backslash \alpha_i)$. It considers the set of elements which are in either of the partitions $\alpha_i$ and $\beta_j$ but not in their intersection (see, e.g., Kuratowski, 2014). Among several properties of this measure, we find that the symmetric difference is commutative and associative. That is, let $\alpha_i$, $\beta_j$ and $\lambda_k$ be three different non-empty partitions of a finite set, then $\alpha_i \Delta \beta_j \Delta \lambda_k = (\alpha_i \Delta \beta_j) \Delta \lambda_k = \alpha_i \Delta (\beta_j \Delta \lambda_k)$. Correspondingly, the Hamming distance between partitions $\alpha_i$ and $\beta_j$ is defined as the cardinality $|\alpha_i \Delta \beta_j|$ of the set $\alpha_i \Delta \beta_j$ (see, e.g., Simovici, 2007). Note that the Hamming distance of two partitions coincides with the Rokhlin distance defined in (1) and consequently in (2). Following a similar analysis, the symmetric difference of three sets or partitions is presented by the

Venn diagram in Fig. 1, where the grey area represents the union of the sets without their successive intersections.

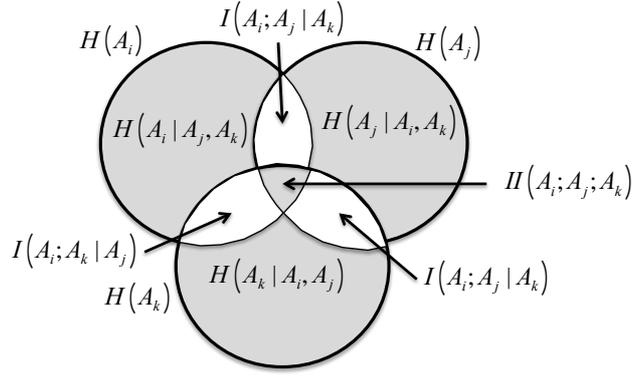

Fig. 1: The Symmetric difference of three non-empty sets is represented by the gray area, together with the information theoretical relationships among the corresponding attributes $A_i$, $A_j$, and $A_k$

Fig. 1 also shows the information theoretical relationships among the attributes $A_i$, $A_j$, and $A_k$, induced by the partitions $\alpha_i$, $\beta_j$ and $\lambda_k$, respectively. Namely, $H(A_i)$, $H(A_j)$, and $H(A_k)$ denote respectively the Shannon entropies of the attributes $A_i$, $A_j$, and $A_k$; while, $H(A_i|A_j)$ for example denotes the conditional entropy of $A_i$ given $A_j$. $I(A_i; A_j)$ is the mutual information between attributes $A_i$ and $A_j$, and $I(A_i; A_j|A_k)$ is the conditional mutual information between attributes $A_i$ and $A_j$. The term $I(A_i; A_j|A_k) = H(A_i|A_k) - H(A_i|A_j, A_k)$ is the conditional mutual information between attributes $A_i$ and $A_j$, given attribute $A_k$ (see Cover and Thomas, 2006). Finally, $II(A_i; A_j; A_k)$ denotes the multivariate mutual information among the three attributes, that was introduced in the seminal work of (McGrill, 1954) as a measure of the higher-order interaction among random variables. Specifically, $II(A_i; A_j; A_k) = I(A_i, A_j; A_k) - I(A_i; A_k) - I(A_j; A_k)$. It can be shown that the multivariate mutual information is symmetric in the three attributes, and that it is bounded from above by $II(A_i; A_j; A_k) \leq \min\{I(A_i; A_j|A_k), I(A_i; A_k|A_j), I(A_j; A_k|A_i),\}$ (McGrill, 1954).

We can now define the measure $d_{MA}$ involving three attributes as follows:

$$d_{MA}(A_i, A_j, A_k) = H(A_i|A_j, A_k) + H(A_j|A_i, A_k) + H(A_k|A_i, A_j) + II(A_i; A_j; A_k), \quad (3)$$

where $II(\cdot)$ is the multivariate mutual information. The first three terms on the right side of (3) represent the degree of uncertainty among attributes, whereas the last term represents the shared information among them.

As an example, consider the dataset $D$ with $p = 4$ and $N = 10$. Table 2 shows the transposed dataset $D$ due to space considerations.

Table 2: An example of a dataset $D$ with $N = 10$ and $p = 4$.

| | | | | | | | | | | |
|---|---|---|---|---|---|---|---|---|---|---|
| $A_1$ | 0 | 1 | 0 | 1 | 0 | 0 | 1 | 0 | 0 | 1 |
| $A_2$ | R | G | R | G | G | R | B | B | R | G |
| $A_3$ | 1 | 2 | 3 | 4 | 5 | 6 | 7 | 8 | 9 | 10 |
| $A_4$ | 1 | 0 | 1 | 0 | 1 | 0 | 0 | 0 | 1 | 0 |

Now, consider the three-attribute distances, $d_{MA}(A_1, A_2, A_3)$, $d_{MA}(A_1, A_3, A_4)$, and $d_{MA}(A_2, A_3, A_4)$. A simple numerical analysis, similar to the one presented in Table 1, implies that $H(A_1) \cong 0.971$; $H(A_2) \cong 1.522$; $H(A_3) \cong 3.322$ and $H(A_4) \cong 0.971$. To compute $d_{MA}(A_1, A_2, A_3)$ we first obtain that

$H(A_1|A_2, A_3) = 0$, $H(A_2|A_1, A_3) = 0$, $H(A_3|A_1, A_2) = 1.276$, and finally $II(A_1; A_2; A_3) = 0.446$. Then, the three-attribute measure $d_{MA}(A_1, A_2, A_3) \cong 1.722$ is obtained. Similarly, the three-attribute measures for the other attributes combinations result in $d_{MA}(A_1, A_2, A_4) \cong 1.469$; $d_{MA}(A_1, A_3, A_4) = 1.722$ and $d_{MA}(A_2, A_3, A_4) \cong 2.771$.

From this simple example, one can understand why the three-attribute set $\{A_1, A_2, A_4\}$ with the lowest multi-attribute measure generates the highest correlative subspace, while the three-attribute set $\{A_2, A_3, A_4\}$ generates the lowest correlative subspace. Let us try to explain the intuition behind it. Note that in Table 2, $A_4$ is highly correlated with $A_1$, hence applying the measure over those two attributes yields the lowest value. Similarly, $A_1$ and $A_2$ are somewhat correlated. On the other hand, $A_3$ has a uniform distribution over its symbols, hence it is not correlated with all the other attributes and thus, potentially, contributes the most to the multi-attribute measure. Taking all the above observations into account, in order to obtain the lowest value for a three-attribute set, which corresponds to the most informative subspace, one needs to neglect $A_3$ and selects $\{A_1, A_2, A_4\}$. On the other hand, the three-attribute set with the highest value, which corresponds to the least informative subspace, has to include $A_3$ and should not include $\{A_1, A_4\}$ nor $\{A_1, A_2\}$, thus resulting in the set $\{A_3, A_2, A_4\}$.

Based on the intuition described above, note the similarly to the Rokhlin distance $d_R$, which is defined in (2) that measures how distant two attributes are. The multi-attribute measure, $d_{MA}$, which is defined in (4), generalizes this quantity to a higher number of attributes. We use this exact property to combine correlated attributes into informative subspaces, when looking for anomalies.

The extension of (3) for $p$ attributes is derived from the definition of the symmetric difference for $p$ sets (see, e.g., Kuratowski, 2014) as follows,

$$d_{MA}(\boldsymbol{A}) = \sum_{i=1}^{p} H(A_i|\boldsymbol{A} \setminus A_i) + II(\boldsymbol{A}), \qquad (4)$$

where $\boldsymbol{A} = \{A_1, A_2, \dots, A_p\}$ denotes a multi-set of attributes in $D$, $A_i \in \boldsymbol{A}$ (note that $\boldsymbol{A}$ can represent a multi-set union of two subsets, where $d_{MA}(\boldsymbol{A_1}, \boldsymbol{A_2}) = d_{MA}(\boldsymbol{A_1} \cup \boldsymbol{A_2}) = d_{MA}(\boldsymbol{A})$ for $\boldsymbol{A} = \boldsymbol{A_1} \cup \boldsymbol{A_2}$); The term $II(\boldsymbol{A})$ is the multivariate mutual information defined for $p > 2$ in (McGrill, 1954). In (Jakulin, 2005) the multivariate mutual information was extended as the recursive computation $II(A_1, A_2, \dots, A_p) = II(A_1, A_2, \dots, A_{p-1}) - II(A_1, A_2, \dots, A_{p-1}|A_p)$. Alternatively, the multivariate mutual information can be defined in measure-theoretic terms as the intersection of individual entropies, i.e., $II(A_1, A_2, \dots, A_p) = H\left(\cap_{i=1}^{p} \tilde{A}_i\right)$, where $\tilde{A}_i$ denotes the abstract set derived from the attribute $A_i$ and $H(\cdot)$ denotes the Shannon entropy (Reza, 1994). The latter definition reflects that the multivariate mutual information is the intersection of all partitions produced by the $p$ attributes. Note that in the case of $p = 2$, the term $II(\cdot)$ is defined as zero. Therefore Eq. (4) reduces to the Rokhlin distance between two partitions.

There are several benefits of using the proposed measure to analyze subspaces as detailed next. First, minimizing the proposed multi-attribute measure corresponds to a selection of informative subspaces that are composed of highly correlated attributes as illustrated in the example in Table 2. Second, unlike classical and state-of-the-art approaches, such as ENCLUS (Cheng et al., 1999), 4S (Nguyen et al., 2014) and CMI (Nguyen et al., 2013), we propose a subspace search algorithm that minimizes the proposed measure instead of maximizing other information measures, such as Total Correlation (Watanabe, 1960). Minimizing the proposed measure is useful for the proposed procedure and it avoids the necessity of selecting a priori some parameters, such as information thresholds, as seen in later sections.

Third, the minimization of the proposed multi-attribute measure tends to delegate the combination of attributes with very low information content or, equivalently, large number of symbols to later stages of the search, where their effects are less critical. To understand this observation, note that low informative attributes often have a higher number of uniformly-alike distributed symbols, e.g., attribute $A_3$ in Table 2. Consequently, the first term in (4) approaches the sum of the Shannon entropy of individual attributes, which by definition yields a higher value than that of the conditional entropy (Cover and Thomas, 2006). Thus, using the proposed method results in a lower value with respect to informative attributes, i.e., attributes with fewer numbers of symbols.

For another illustration, consider Table 2 and compute the Total Correlation (TC) according to (Watanabe, 1960), $TC(A_1, A_2, ..., A_p) = \sum_{i=1}^{p} H(A_i) - H(A_1, A_2, ..., A_p)$. Conceptually, TC quantifies the amount of information that is shared among the different attributes and thus expresses how the attributes are related to each other. The values of $TC$ for all the possible combinations of three attributes in Table 2 are the following: $TC(A_1, A_2, A_3) \cong 2.493$ ; $TC(A_1, A_2, A_4) \cong 1.093$ ; $TC(A_1, A_3, A_4) \cong 2.493$; and $TC(A_2, A_3, A_4) \cong 2.493$. Thus, the TC implies selecting either one of the subspaces $\{A_1, A_2, A_3\}$, $\{A_1, A_3, A_4\}$, or $\{A_2, A_3, A_4\}$ without distinguishing between them. Differently from TC, the multi-attribute measure correctly indicates that the subspace $\{A_1, A_2, A_4\}$ is the most informative one. Since $A_3$ is not selected, the latter subspace represents a higher qualitative subspace when seeking for anomaly detection applications. We argue that $TC$ captures a different measure of information on the subspaces than the proposed multi-attribute distance. Moreover, since $TC$ is a non-decreasing function, it is maximized when all the attributes are selected. Thus, in contrast to the proposed $d_{MA}$, the use of $TC$ for selecting informative subspaces requires to decide in advance on the number of attributes in a subspace, leading potentially to a selection of uninformative subspaces and suboptimal solutions like other parametric methods.

Finally, later sections empirically show that minimization of the proposed measure tends to generate, on average, a smaller set of subspaces than other approaches, especially in the case of datasets whose attributes have a considerably high number of unique values. A direct consequence of this characteristic is a reduced training time, on average, of the ensemble models.

### 3.3 Approximation of the Multi-Attribute Distance for p > 2

As $p$ grows, the probability distributions are becoming higher-dimensional, and hence the estimation of the multi-attribute measure becomes less reliable. To address this issue, we make use of the following claim on $d_{MA}$ given two sets of attributes $A_i$ and $A_j$:

**Lemma 1.** $A_j \subseteq A_i \Rightarrow d_{MA}(A_j) \geq d_{MA}(A_i)$.
**Proof:** Refer to Appendix B.1.

An immediate result of Lemma 1 is the following: given a set of subsets of $A$ denoted by $\widetilde{A}$, $d_{MA}(A) \leq \min_{A_j \in \widetilde{A}}\{d_{MA}(A_j)\}$. In other words, $\min_{A_j \in \widetilde{A}}\{d_{MA}(A_j)\}$ can serve as an upper bound approximation of $d_{MA}(A)$.

Therefore, when the size of $\widetilde{A}$ is relatively small (say all subsets of $A$ are of size 2 or 3 attributes), then

$d_{MA}(A)$ can often be approximated efficiently. The approximated $d_{MA}$ can be applied to compute the information "distance" within a subspace, and consequently can be used in the search for highly correlative subspaces, as shown in section 5. In our experiments, we found that the use of more than three attributes compromises the computation of the conditional entropies in $d_{MA}(\cdot)$. Specifically, when having a large number of attributes, the most refined partition often resulted in single-value elements in each of its subsets.

## 4 Agglomerative Attribute Grouping

In this section, we present our subspace analysis method, which is named the Agglomerative Attritute Grouping (AAG). Similar to the subspace analysis methods described in section 2.2, AAG generates a set of subspaces with highly correlated attributes by applying a variation of the well-known agglomerative clustering algorithm and using the proposed $d_{MA}$ measure as the underlying "distance" function. Combining this measure with the agglomerative strategy, can be used to find subspace combinations without the need to set any parameter value in advance (including the number of subspaces we are looking for). This is one of the major differences in comparison to other conventional methods, e.g., ENCLUS (Cheng et al., 1999), FB (Lazarevic and Kumar, 2005), HiCS (Keller et al., 2012), CMI (Nguyen et al., 2013) and 4S (Nguyen et al., 2014).

The pseudocode of the proposed AAG method is shown in Algorithm 1. The algorithm receives as input a dataset $D$ composed of $N$ observations and $p$ attributes. The algorithm returns as output a set of subspaces with highly correlated attributes denoted by $T$. The algorithm begins by initializing the result set of subspaces $T$ to be the empty set (line 1). Then, in line 2, the algorithm generates a set of $n$ subspaces, each of which is composed of a single attribute. This set constitutes the first agglomeration level and is denoted by $S^{(t)}$, $t = 1$ (lines 2-3). Then, the algorithm iteratively generates the subspaces of agglomeration level $t + 1$, denoted by $S^{(t+1)}$ by combining subspaces from the previous agglomeration level, $S^{(t)}$ (lines 4-27). Each such iteration begins with updating the result set $T$ to contain also the subspaces from the previous agglomeration level (line 5). Then, in line 6, we initialize the set of subspaces of the next agglomeration level to be the empty set. Next, in line 7, we maintain a copy of the previous agglomeration level, denoted by $S_0^{(t)}$. This is required to allow attributes to appear in different subspaces. Notice that $S_0^{(t)}$, $S^{(t)}$ and $S^{(t+1)}$, as well as $T$, contain the indices of the data attributes in the subspaces, whereas, e.g., $A_i$ denotes the projection of data samples.

The algorithm continues by searching for two subsets in the current agglomeration level that have the lowest $d_{MA}$ value (line 8) and adds the unified set to the next agglomeration level instead of the two individual subsets (lines 9). In lines 10-12 (and also later in lines 18-20), the algorithm can choose not to add the resulting set; we refer to this stage as the pruning stage and describe it in detail in subsection 4.1. In lines 13-25, the algorithm continues to combine subspaces iteratively until there are no more subsets left in $S^{(t)}$. However, now, the algorithm checks whether it is better to unify a subset from $S^{(t)}$ and a subset from $S^{(t+1)}$, denoted by $A_i$ and $A_j$, or two subsets from $S^{(t)2}$, denoted by $A_i$ and $A_k$. The motivation behind this stage is to avoid merging only a single pair of subspaces in each agglomeration level and to allow the merging of multiple subspaces. In doing so, we avoid the permanent selection of subspaces with a higher number of attributes to be combined.

---
[2] Note that in order to reduce runtime complexity considerably, we do not iterate over all pairs of subsets in $S^{(t)}$, but only on pairs that include $A_i$ and another subset (i.e., $A_k$) from $S_0$.

Once all subspaces have been assigned at agglomeration level $t$, the algorithm proceeds with subsequent levels of agglomeration (lines 4-27) until no subspace combination is further required (line 4). The AAG algorithm ends by returning the set of subspaces $T$ in line 28.

The normalized multi-attribute measure, denoted by $\tilde{d}(\cdot)$ as used in lines 8, 14, 15 and 16, is defined in (5).

$$\tilde{d}(A_i, A_j) = \frac{d_{MA}(A_i, A_j)}{H(A_i \cup A_j)} \quad (5)$$

where $d_{MA}(A_i, A_j)$ is defined in (4) and $H(A_i \cup A_j)$ denotes the join entropy after unifying the subspaces $A_i$ and $A_j$. The normalization factor in (5), i.e. $H(A_i \cup A_j)$, allows a comparison between subspaces with different numbers of attributes. We used the results from (Yianilos, 2002) that showed that this normalization factor does not change the measure characteristics of (4). In the general case, the computation of the measure is obtained based on Lemma 1 where we select a fixed-size number of attributes (e.g., three), and calculate the minimum value over all subsets of this given size.

**Algorithm 1: Agglomerative Attribute Grouping**

**Input:** A data set $D$ with $N$ observations and $p$ attributes
**Output:** A set of subspaces in $T$

1:    $T \leftarrow \emptyset$
2:    $S^{(1)} \leftarrow \{\{A_1\}, \{A_2\}, \dots, \{A_p\}\}$
3:    $t \leftarrow 1$
4:    **while** $(S^{(t)} \neq \emptyset)$ **do**
5:      $T \leftarrow T \cup S^{(t)}$
6      $S^{(t+1)} \leftarrow \emptyset$
7:      $S_0 \leftarrow S^{(t)}$
8:      $\{A_i, A_j\} = \underset{A_i, A_j \in S^{(t)}}{\operatorname{argmin}} \tilde{d}(A_i, A_j)$
9:      $S^{(t)} \leftarrow S^{(t)} \setminus \{A_i, A_j\}$
10:     **if** $(t \leq 2)$ $OR$ $(TC(A_i \cup A_j) \geq v_i TC(A_i) + v_j TC(A_j))$ **then**
11:      $S^{(t+1)} \leftarrow S^{(t+1)} \cup \{A_i \cup A_j\}$
12:     **end if**
13:     **while** $(S^{(t)} \neq \emptyset)$ **do**
14:      $\{A_i, A_j\} = \underset{A_i \in S^{(t)}, A_j \in S^{(t+1)}}{\operatorname{argmin}} \tilde{d}(A_i, A_j)$
15:      $A_k = \underset{A_k \in S_0 \setminus A_i}{\operatorname{argmin}} \tilde{d}(A_k, A_i)$
16:      **if** $\left(\tilde{d}(A_i, A_j) \geq \tilde{d}(A_i, A_k)\right)$ **then**
17:       $S^{(t)} \leftarrow S^{(t)} \setminus \{A_i, A_k\}$
18:      **if** $(t \leq 2)$ $OR$ $(TC(A_i \cup A_j) \geq v_i TC(A_i) + v_j TC(A_j))$ **then**
19:       $S^{(t+1)} \leftarrow S^{(t+1)} \cup \{A_i \cup A_k\}$
20:      **end if**
21:      **Else**
22:       $S^{(t)} \leftarrow S^{(t)} \setminus A_i$
23:       $A_j \leftarrow A_i \cup A_j$
24:      **end if**
25:     **end while**
26:     $t \leftarrow t + 1$

```
27:     end while
28:     return T
```

To illustrate the operation of the proposed algorithm, consider a dataset $D$, with $p = 7$ attributes as shown in Table 3.

Table 3: A dataset $D$ where $N = 10$ and $p = 7$ are used in the running example to illustrate the operation of the proposed AAG algorithm.

| $A_1$ | $A_2$ | $A_3$ | $A_4$ | $A_5$ | $A_6$ | $A_7$ |
|---|---|---|---|---|---|---|
| 0 | R | 1 | 1 | a | 3 | 9 |
| 1 | G | 2 | 0 | a | 3 | 9 |
| 0 | R | 3 | 1 | a | 5 | 25 |
| 1 | G | 4 | 0 | a | 5 | 25 |
| 0 | G | 5 | 1 | a | 7 | 49 |
| 0 | R | 6 | 0 | b | 8 | 64 |
| 1 | B | 7 | 1 | b | 10 | 100 |
| 0 | B | 8 | 0 | b | 10 | 100 |
| 0 | R | 9 | 0 | b | 11 | 121 |
| 1 | G | 10 | 0 | a | 11 | 121 |

The AAG method starts by initializing the set of subspaces of level 1, denoted by $S^{(1)} = \{\{A_1\}, \{A_2\}, \{A_3\}, \{A_4\}, \{A_5\}, \{A_6\}, \{A_7\}\}$ (line 2). Then, AAG searches for a pair of subspaces in $S^{(1)}$ that minimizes $\tilde{d}(\cdot)$ (line 8). Going over all the 21 possible pairs and using (4), we find that the pair $\{A_6\}$ and $\{A_7\}$ minimizes this measure where $\tilde{d}(\{A_6\}, \{A_7\}) \cong 0$. Then, the two subspaces $\{A_6\}$ and $\{A_7\}$ are removed from $S^{(1)}$ (line 9), and the unified subspace $\{A_6, A_7\}$ is added to the set of subspaces of level 2, denoted by $S^{(2)}$ (line 11). At this point $S^{(1)} = \{\{A_1\}, \{A_2\}, \{A_3\}, \{A_4\}, \{A_5\}\}$ and $S^{(2)} = \{\{A_6, A_7\}\}$. Next, AAG keeps searching for higher-order subspaces by combining subspaces from $S^{(1)}$ with subspaces from $S^{(2)}$ (line 14). Going over all five possible pairs (there is only one subspace in $S^{(2)}$ and five subspaces in $S^{(1)}$), it is found that the pair $\{A_6, A_7\}$ and $\{A_1\}$ minimizes the measure, where $\tilde{d}(\{A_6, A_7\}, \{A_1\}) \cong 0.292$. Next, the algorithm checks whether $\{A_1\}$ is more informative w.r.t. $\tilde{d}$ to other subspaces in $S_0^{(1)}$ than to $\{A_6, A_7\}$ (line 15).

As a practical note, notice that all the computations involving $\{A_1\}$ and other subspaces in $S_0^{(1)}$ have already been computed in the previous iteration when $\{A_6, A_7\}$ was chosen. Such computations can be stored in a look-up table and reduce future computations considerably. We find that the subspace in $S_0^{(1)}$ that minimizes $\tilde{d}$ is $\{A_1\}$ with $\tilde{d}(\{A_1\}, \{A_3\}) \cong 0.708$. Since $\tilde{d}(\{A_1\}, \{A_3\}) > \tilde{d}(\{A_6, A_7\}, \{A_1\})$, $\{A_1\}$ is combined with $\{A_6, A_7\}$, yielding the new subspace $\{A_1, A_6, A_7\}$ (line 16). Then, $\{A_1\}$ is removed from $S^{(1)}$ (line 22), and the new subspace $\{A_6, A_7, A_1\}$ replaces the subspace $\{A_6, A_7\}$ in $S^{(2)}$. Thus, $S^{(1)} = \{\{A_2\}, \{A_3\}, \{A_4\}, \{A_5\}\}$ and $S^{(2)} = \{\{A_1, A_6, A_7\}\}$. Next, the algorithm proceeds to search for a subspace in $S^{(1)}$ that, if combined with $\{A_1, A_6, A_7\}$, will keep it highly informative. Since the combined subspaces now contain four attributes, when computing $\tilde{d}$, we apply Lemma 1 and select only three attributes. More specifically, we compute $\tilde{d}(\{A_1, A_6, A_7\}, \{A_2\})$, $\tilde{d}(\{A_1, A_6, A_7\}, \{A_3\})$, $\tilde{d}(\{A_1, A_6, A_7\}, \{A_4\})$, and $\tilde{d}(\{A_1, A_6, A_7\}, \{A_5\})$ and find that $\tilde{d}(\{A_1, A_6, A_7\}, \{A_3\}) \cong 0.051$ minimizes $\tilde{d}$. The subspace $\{A_1, A_6, A_7\}$ is therefore replaced with $\{A_1, A_3, A_6, A_7\}$, and $\{A_3\}$ is removed form $S^{(1)}$ (line 22), yielding $S^{(1)} = \{\{A_2\}, \{A_4\}, \{A_5\}\}$ and $S^{(2)} = \{\{A_1, A_3, A_6, A_7\}\}$. Since $S^{(1)} \neq \emptyset$, the algorithm continues (line 13) to search for combinations of subspaces from $S^{(1)}$ and $S^{(2)}$ that minimize $\tilde{d}(\cdot)$ (lines 14-25). The AAG method finds that combining $\{A_1, A_3, A_6, A_7\}$ and $\{A_4\}$ yields the minimum value with

$\tilde{d}(\{A_1, A_3, A_6, A_7\}, \{A_4\}) \cong 0.443$. However, in line 15, it finds that $\tilde{d}(\{A_2\}, \{A_4\}) < \tilde{d}(\{A_1, A_3, A_6, A_7\}, \{A_4\})$, and therefore, it unifies $\{A_2\}$ and $\{A_4\}$ into $\{A_2, A_4\}$, adds it to $S^{(2)}$ and removes the latter two single-attribute subspaces from $S^{(1)}$. At this point $S^{(2)} = \{\{A_1, A_3, A_6, A_7\}, \{A_2, A_4\}\}$ and $S^{(1)} = \{\{A_5\}\}$.

Similarly, since $\tilde{d}(\{A_1, A_3, A_6, A_7\}, \{A_5\}) < \tilde{d}(\{A_2, A_4\}, \{A_5\})$ (lines 14-15), $\{A_5\}$ is unified with $\{A_1, A_3, A_6, A_7\}$, replacing the latter one in $S^{(2)}$. This stage results in $S^{(2)} = \{\{A_1, A_3, A_5, A_6, A_7\}, \{A_2, A_4\}\}$, and $S^{(1)} = \emptyset$, unsatisfying the condition in line 13 and breaking the loop. Then, the next level adds the union of the two subspaces in $S^{(2)}$ to $S^{(3)}$ leaving $S^{(2)}$ empty and breaking the loop in line 13. Finally, the algorithm terminates and returns $T = \{\{A_1, A_3, A_5, A_6, A_7\}, \{A_2, A_4\}, \{A_1, A_2, A_3, A_4, A_5, A_6, A_7\}\}$. Fig. 2 shows the resulting subspace combination in form of an agglomerative tree. Note that $\mathbf{A}_1 = \{A_1, A_3, A_5, A_6, A_7\}$, $\mathbf{A}_2 = \{A_2, A_4\}$ and $\mathbf{A}_3 = \mathbf{A}_1 \cup \mathbf{A}_2 = \{A_1, A_2, A_3, A_4, A_5, A_6, A_7\}$. Note also that, in this particular example, each attribute was assigned to exactly one subspace in each one of the agglomeration levels; however, due to line 14 in the algorithm, this is not necessarily always the case.

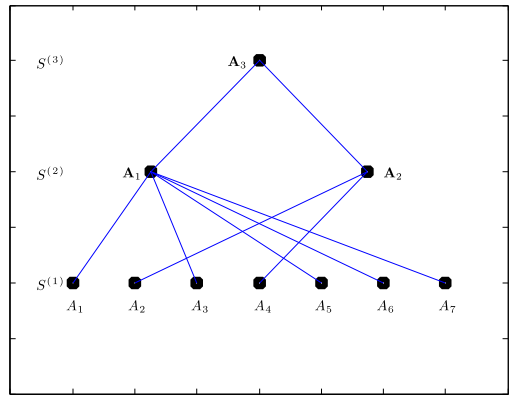

Fig. 2: Resulting agglomeration tree from the illustrative running example.

## 4.1 Pruning Stage

The agglomerative approach used in the previous section has an inherent property that the number of attributes in subspaces grows with the agglomeration level. This property has two major limitations: (i) it may have a great impact on the efficiency of the anomaly detection ensemble and, (ii) recall that (4) becomes less accurate when the number of attributes grows considerably. To overcome these limitations, we propose a simple rule to determine whether to proceed with unifying two subspaces or not. This rule is embedded in the algorithm shown in Algorithm 1 in lines 10-12 and 18-20. According to this rule, two candidate subspaces are unified only if their union does not considerably reduce the subspace's quality with respect to the two individual subspace candidates. More specifically, we evaluate the Total Correlation (TC) (Watanabe 1960) of the two individual subspaces $\mathbf{A}_i$ and $\mathbf{A}_j$ and compare their sum to the TC of their union $\mathbf{A}_i \cup \mathbf{A}_j$:

$$TC(\mathbf{A}_i \cup \mathbf{A}_j) \geq v_i TC(\mathbf{A}_i) + v_j TC(\mathbf{A}_j) \qquad (6)$$

where $v_i = J(\mathbf{A}_i; \mathbf{A}_i \cup \mathbf{A}_j)$ and $v_j = J(\mathbf{A}_j; \mathbf{A}_i \cup \mathbf{A}_j)$ serve as soft thresholds and $J(\cdot)$ is the well-known Jaccard Index. If the condition is satisfied (the sum of individual $TC$s is lower than the $TC$ of their union), the two subspaces are combined. Note that the proposed rule does not require any tuning of parameters. Moreover, its usage by AAG does not lead to discarded attributes since all attributes are already combined in the previous level of agglomeration. As noted above, this is an important property in anomaly detection applications where all attributes are required.

In some special cases, it is possible to speed up the evaluation of the rule by avoiding the computation of the different $TC$s. For example, if $A_i \cap A_j = \emptyset$, the following Lemma indicates that it is legitimate to unify the two subspaces:

**Lemma 2:** *Given two subspaces $A_i$ and $A_j$, such that $|A_i| \geq 2$ and $|A_j| \geq 2$, and $A_i \cap A_j = \emptyset$, then necessarily $TC(A_i \cup A_j) \geq TC(A_i) + TC(A_j)$.*

**Proof**: Refer to Appendix B.2.

Note that, for $A_i \cap A_j = \emptyset$, the soft thresholds result in $v_i \leq \delta$ and $v_j \leq 1 - \delta$, where $\delta \in (0, 0.5)$ is to be computed. Furthermore, it can be shown that, if $A_i \subseteq A_j$ (or $A_j \subseteq A_i$), then, $TC(A_i \cup A_j) = TC(A_j) \leq v_i TC(A_i) + TC(A_j)$ (since $v_j = 1$, and $v_i > 0$). Therefore, in such cases, the two subspaces should not be unified.

Also note that while $TC$ is not a formal metric, it can still be used for comparison (i.e., testing whether one set is "better" than the other), as implemented in Equation (6).

## 4.2 Complexity Analysis

In this subsection, we analyze the complexity of Algorithm 1. Note that, since we focus on the worst-case scenario, the pruning mechanism is ignored.

In line 8 of Algorithm 1, AAG searches for the two subspaces with minimal $\tilde{d}(\cdot)$, among all possible combinations of single-attribute subspaces. Because we have $p$ attributes in total, the runtime complexity of line 8 is $O(p^2 \Delta)$, where $\Delta$ represents the time required to compute $\tilde{d}(\cdot)$. Although the algorithm searches only for the pair of subspaces with minimal $\tilde{d}$, the $\tilde{d}$ values between all pairs are recorded in a matrix $M$. Since $\tilde{d}(\cdot)$ is symmetric, only $p(p-1)/2$ computations are executed and stored. The search for successive subspaces occurs in lines 14 and 15, where subspaces at levels $S^{(t)}$ and $S^{(t+1)}$ are compared. The runtime complexity of line 14 is again $O(p^2 \Delta)$. In line 15, the algorithm uses the previously computed matrix $M$. Thus, only a linear search with one comparison is required, resulting in a runtime complexity of $O(p)$.

When the algorithm reaches line 26, it has combined $k \leq p/2$ subspaces, and a new iteration begins in line 8. That is, in the worst-case scenario of the AAG algorithm, the number of subspaces to be analyzed is reduced by half. Therefore, the maximal number of iterations is $O(\log p)$.

The computation of $\Delta$ requires the estimation of the conditional entropy among attributes, as well as the multi-variate mutual information in $\tilde{d}(\cdot)$. To analyze the complexity $\Delta$ we start by first assuming that $\tilde{d}(\cdot)$ is applied to two attributes. In this case, one attribute out of $p$, e.g., $A_i$, partitions the dataset $D$ by identifying its $m_i$ unique values. The run time to find unique $m_i$ elements in an array of size $N$ can be estimated by $O(N)$ (e.g., by iterating over the elements in the list and adding them into a hash-set). Following this, the unique $m_j$ elements of the second attribute, e.g., $A_j$, at each one of the $m_i$ partitions are identified. This step requires again a running time of $O(N)$ (iterating over the elements of each partition separately, and doing so for all partitions, is equivalent to iterating once over all elements of the list). The computation of the normalization factor $H(A_i, A_j)$ in $\tilde{d}(\cdot)$ does not require additional computation since the values $H(A_i, A_j) = H(A_i) + H(A_j|A_i)$ (see, e.g., Cover and Thomas, 2006) are already computed. Thus, the runtime complexity $\Delta$ can be estimated as $O(2N)$ for two attributes. Next, for three attributes in $\tilde{d}(\cdot)$, $\Delta$ can be estimated as $O(3N)$. Notice that the multi-variate mutual information $II(\cdot)$ for three attributes does not require new computations since, $II(A_i; A_j; A_k) = I(A_i; A_j) - I(A_i; A_j|A_k)$. Furthermore, $I(A_i; A_j) = H(A_i) - H(A_i|A_j)$ and $I(A_i; A_j|A_k) = H(A_i|A_k) - H(A_i|A_j, A_k)$. Finally, the normalization factor in $\tilde{d}(\cdot)$ for three attributes is computed using the chain rule, i.e., $H(A_i, A_j, A_k) = H(A_i) + H(A_i|A_j) + H(A_i|A_j, A_k)$ (see, e.g., Cover and Thomas, 2006). That is, all the later terms can be computed using previously assessed values.

Combining the analysis done for both the agglomerative strategy of AAG and the computation of $\tilde{d}(\cdot)$, the runtime complexity of AAG can finally be estimated as $O(Np^3 \log p)$.

Note that the pruning rule can be approximately computed in $O(pN)$ as we further explain. Assuming a subspace set $S_i$ comprised of $k$ attributes $\{A_1, A_2, \ldots, A_k\}$, then the Total Correlation $TC(A_1, A_2, \ldots, A_k) = \sum_{j=1}^{k} H(A_j) - H(A_1, A_2, \ldots, A_k) = \sum_{j=1}^{k} H(A_j) - \sum_{j=1}^{k} H(A_j|A_{j-1}, \ldots, A_1)$. Now, let $A_m$ be the attribute with the maximum Shannon entropy $H(A_m)$ for the attributes in the subspace $S_i$. Since conditioning cannot increase entropy, $\sum_{j=1}^{k} H(A_j|A_{j-1}, \ldots, A_1) \leq \sum_{j=1}^{k} H(A_j|A_{j-1})$, and in the case that attributes are maximally related to each other, $\sum_{j=1}^{k} H(A_j|A_{j-1}) \leq H(A_m)$. Thus, $TC(A_1, A_2, \ldots, A_k) \leq \sum_{j=1}^{k} H(A_j) - H(A_m)$. Therefore, the Shannon entropy of each attribute can be pre-computed during the initialization phase of the AAG algorithm with a runtime complexity of $O(pN)$. These values are then stored into an array of length $p$. Therefore, the runtime complexity at lines 10 and 18 in Algorithm 1 can be computed as $O(p)$. Following this analysis, the runtime complexity of AAG remains unmodified.

## 5 Evaluation

In this section, we compare the quality of the subspaces generated by AAG against eight other benchmark algorithms, when used in ensembles for anomaly and novelty detection.

### 5.1 Experimental Settings

Our empirical study is based on the experimental settings used in (Keller et al., 2012; Cheng et al., 1999; and Nguyen et al., 2013). All of our experiments were conducted on 25 real-world datasets (see Table 4) taken from the UCI repository (Bache and Lichman, 2013). Although these datasets are usually used in the context of classification tasks, previous studies (e.g., Aggarwal and Yu, 2001; Lazarevic and Kumar, 2005; Keller et al., 2012; Nguyen et al., 2013; and Nguyen et al., 2014) have also used them in the context of anomaly and novelty detection. In subsection 5.1.1, we describe in details how normal and abnormal observations for each dataset were generated, and how the training and test sets were obtained. Specifically, we consider three different settings (two of them are related to anomaly detection and the third is related to novelty detection).

The following evaluation procedure was used for AAG as well as for the eight benchmark algorithms (see section 5.1.2). Stated differently, the only difference between the evaluation procedure of the various methods was the subspace analysis algorithm used.

First, each subspace analysis algorithm was learnt over the training set. Then, the same training set was used to train the anomaly detection algorithm (we used MV-Set, more details are provided in section 5.1.3) in each one of the obtained subspaces.

The missing values in each attribute of the training dataset were replaced by the mean value in case of non-categorical valued attributes, and by the most frequent symbol in case of categorical-values attributes (see, e.g., Bishop, 2006). The missing values in the test dataset were accordingly replaced by the mean and most frequent values computed from the training dataset. Since AAG, ENCLUS, and 4S make use of elements of information theory to combine subspaces, we discretized the continuous-valued attributes in the training set using the *Equally Frequency* technique following the recommendations by Garcia et al. (2013).

After training the anomaly detection algorithm over each subspace, a weighting factor was computed to aggregate the ensemble elements at the test stage. To this purpose, we followed the recommendations of Menahem et al., (2013). More specifically, the training data was split randomly into a new training dataset, which was used to generate the subspaces as well as to train the MV-Set model in each subspace, and into a validation dataset, which was used to estimate the generalization error of each trained model. That is, the validation data (i.e., majority class) was used to compute the weighting factors as the average error of the MV-Set in each subspace to be used as a "belief factor" of how good each trained model represents the normal data in each subspace. Note that since the validation data contains only normal observations, only one type of error is considered (i.e., normal observations that were classified as anomalies). The aggregation of the ensemble elements was incorporated by summing up the weighted factors of the subspaces as follows. Given an observation $x$ from the validation dataset we computed $\hat{y} = \sum_{i=1}^{M} w_i g_i(x) \geq \rho$, where $\hat{y} \in \{0,1\}$ denotes if the observation is normal (i.e., $\hat{y} = 1$); $w_i$ denotes the weighting factor of subspace $i = 1,2,\ldots,|T|$; $|T|$ denotes the total number of subspaces; $g_i(x)$ represents the MV-Set model trained on subspace $i$, and $\rho$ denotes a threshold computed as the weighted number of subspaces that guaranteed at maximum $\alpha$ error rate on the validation dataset. As default we used $\alpha = 0.05$, as it is typically used in many academic and industrial applications. It is important to emphasize that, in all of our experiments, we only used the normal observations to find subspaces and to train the ensemble for anomaly detection, since only this information is assumed to be available at the training stage. In other words, our training set did not contain any abnormal observations at all.

Finally, the trained ensemble for anomaly detection was evaluated over the test set (containing both normal and abnormal observations).

As a measure of performance, we computed the F1-Score which is calculated as F1= 2TP / (2TP + FN + FP), where TP, FP and FN denote, respectively, the number of True Positives (true anomaly samples), the number of False Positives (number of normal samples classified as anomalies), and the number of False Negatives (the number of anomalies classified as normal samples).

All experiments were executed 20 times, where in each repetition, the dataset was re-split randomly into training and test sets. The reported results are averages over the 20 different repetitions.

Finally, we also tested the statistical significance of the results by applying the evaluation methodology recommended in (Demšar, 2006). Specifically, we first applied the non-parametric Friedman method to test the null hypothesis that AAG is indifferent from its competitors. As recommended by Demšar, (2006), we used the Iman-Davenport correction to generate a statistical value that follows an F-distribution. Then, in cases where the null hypothesis was rejected, we performed the post hoc Bonferroni-Dunn statistical test between AAG and each one of the benchmark methods.

All of the experiments were conducted on a standard MacBook Pro running Mac OS X Version 10.6.8, with a 2.53-GHz Intel® Core 2 Duo processor and 8 GB of DRAM.

Table 4: Characteristics of 25 UCI public datasets used in this study.

| Dataset | Instances | Dimensionality |
|---|---|---|
| KDD Cup 99 (HTTP) | 567479 | 3 |
| KDD Cup 99 (SMTP) | 95156 | 3 |
| Thyroid | 3772 | 6 |
| Mammography | 11183 | 6 |
| Glass | 214 | 9 |
| Breast Cancer | 683 | 9 |
| Zoo | 101 | 10 |
| Cover | 286048 | 10 |
| Wine | 129 | 13 |
| Pen-Digits | 6870 | 16 |
| Letter | 12551 | 16 |
| Waveform 1 | 1826 | 21 |
| Faults | 804 | 27 |
| Dermatology | 194 | 34 |
| Satimage | 5803 | 36 |
| Waveform 2 | 1860 | 40 |
| Segmentation | 4410 | 50 |
| Lung Cancer | 32 | 56 |
| Sonar | 117 | 60 |
| Features Pix | 1244 | 64 |
| Audiology | 226 | 69 |
| Feature Fourier | 866 | 76 |
| MNIST | 7603 | 100 |
| Features Kar | 1244 | 240 |
| Arrhythmia | 292 | 279 |

### 5.1.1 The Three Considered Settings

As explained above, we considered three different settings, as detailed next.

**Setting 1 – Anomaly Detection (Adding Gaussian Noise):** In this setting we simulated a case where anomalies were generated by adding zero-mean Gaussian noise to normal observations, but only over a subset of the attributes and not over the entire data space. More specifically, we first identified the majority class for each one of the datasets. Then, we sampled 70% of the observations associated with the majority class. These observations were considered as normal observations and served as the training set. The remaining 30% of the observations associated with the majority class were split into two equally sized datasets. One of the newly split sets was kept as is, representing normal observations in the test set. For the other split, we randomly selected $K$ attributes from the entire data space and added zero-mean Gaussian noise on the projected subspace of these attributes, representing anomalies in the test set. The Variance-Covariance matrix of the Gaussian noise was set to be diagonal, where the diagonal elements are the variances of the $K$ attributes in the selected subspace. The described procedure was repeated with different percentages of perturbed attributes, i.e. 1%, 3%, 5%, 7%, and 10% to 100% with steps of 10%.

**Setting 2 – Anomaly Detection (Merging Classes):** In this setting, we applied the approach used by Emmott et al. (2013). For datasets with binary classes, we arbitrarily selected one class as normal and the other one as an anomalous class. For the cases of multi-class datasets, we trained a Random Forest classifier and based on the resulting confusion matrix, we identified pairs of classes that were commonly confused against each other and merged them into an anomalous class. The rest of the classes were merged into the normal class. Finally, the training set was composed of 70% of the observations associated with the normal class, and the test set was composed of the remaining 30%, as well as 5% of the observations associated with the anomalous class. Note that we follow the terminology used by Emmott et al. (2013) and relate to this setting as anomaly detection, but it is important to note that this setting can also be considered as novelty detection.

**Setting 3 – Novelty Detection:** In this setting, we simulated a case where the abnormal observations represent a previously unseen class, i.e., novelties as defined in the literature. For this purpose, we used the approach that was applied in several previous studies (see, e.g., Aggarwal and Yu, 2001; Lazarevic and Kumar, 2005; Keller et al., 2012; Nguyen et al., 2013; and Nguyen et al., 2014). Similar to the first setting described above, we first sampled 70% of the observations associated with the majority class. These observations represented normal observations and served as the training set. The remaining 30% of the observations associated with the majority class represented normal observations in the test set. Finally, 10% of the observations associated with the remaining classes (i.e., not with the majority class) represented novelties in the test.

To summarize this section, Table 5 shows the number of normal and abnormal instances for each one of the datasets for each one of the settings.

Table 5: Characteristics of 25 UCI public datasets used in this study.

| Dataset | Setting 1 | | Setting 2 | | Setting 3 | |
|---|---|---|---|---|---|---|
| | Normal Instances | Abnormal Instances | Normal Instances | Abnormal Instances | Normal Instances | Abnormal Instances |
| KDD Cup 99 (HTTP) | 565287 | 2211 | 480494 | 84793 | 565287 | 2211 |
| KDD Cup 99 (SMTP) | 95126 | 30 | 80857 | 14269 | 95126 | 30 |
| Thyroid | 3679 | 93 | 3127 | 552 | 3679 | 93 |
| Mammography | 10923 | 260 | 9285 | 1638 | 10923 | 260 |
| Glass | 205 | 9 | 174 | 31 | 76 | 9 |
| Breast Cancer | 444 | 239 | 377 | 67 | 444 | 239 |
| Zoo | 41 | 6 | 35 | 6 | 41 | 6 |
| Cover | 283301 | 2747 | 240806 | 42495 | 283301 | 2747 |
| Wine | 119 | 10 | 101 | 18 | 71 | 6 |
| Pen-Digits | 6714 | 156 | 5707 | 1007 | 780 | 85 |
| Letter | 8047 | 100 | 6840 | 1207 | 8047 | 813 |
| Waveform 1 | 1657 | 169 | 1408 | 249 | 1696 | 171 |
| Faults | 739 | 65 | 628 | 111 | 673 | 70 |
| Dermatology | 181 | 13 | 154 | 27 | 112 | 14 |
| Satimage | 5732 | 71 | 4872 | 860 | 1342 | 137 |
| Waveform 2 | 1692 | 168 | 1438 | 254 | 1692 | 170 |
| Segmentation | 112 | 14 | 95 | 17 | 30 | 12 |
| Lung Cancer | 23 | 9 | 20 | 3 | 23 | 9 |
| Sonar | 111 | 12 | 94 | 17 | 111 | 12 |
| Features Pix | 1200 | 44 | 1020 | 180 | 200 | 27 |
| Audiology | 57 | 27 | 48 | 9 | 57 | 27 |
| Feature Fourier | 1200 | 44 | 1020 | 180 | 200 | 27 |
| MNIST | 6903 | 700 | 5868 | 1035 | 6903 | 700 |
| Features Kar | 1200 | 44 | 1020 | 180 | 200 | 27 |
| Arrhythmia | 386 | 66 | 328 | 58 | 245 | 30 |

### 5.1.2 Benchmark Algorithms for Subspace Analysis

As benchmark methods against the proposed AAG method, we selected eight classical and state-of-the-art algorithms, representing a wide range of techniques. Specifically, FB (Lazarevic and Kumar, 2005) and Isolation Forest (Lie, Thing et al., 2008) were selected to represent the random selection of attributes. HiCS was selected to represent the a-priori based technique (Keller et al., 2012). ENCLUS (Cheng et al., 1999), EWKM (Jing et al., 2007) and AFG-*k*-means (Gan and Ng, 2015) were selected to represent the clustering-based techniques. Finally, CMI (Nguyen et al., 2013) and 4S (Nguyen et al., 2014) were selected to represent a category of algorithms that search for subspaces based on information theoretical measures.

With regard to AAG, three attributes were used in the evaluation of (4), which seemed to be a good trade-off between high-quality subspaces and a reasonable run-time. Our implementation of FB sampled attributes from a uniform distribution over the range $[p/2, p]$ as suggested in (Lazarevic and Kumar, 2005). The total number of subspaces (i.e., ensemble size) was set to 20 according to the authors' suggestion.

Our implementation of the Isolation Forest (iForest) algorithm tightly followed the work published in (Liu et al., 2008). iForest generates and ensembles decision trees, where attributes and splits are randomly selected. Each tree, denoted as isolation tree, is built recursively by partitioning the given feature space until samples are isolated. The height (node-depth) of each sample is mapped to a score. Normal samples are then expected to be associated leaves of average height, whereas abnormal samples are expected to be associated leaves with lower height.

The HiCS algorithm was executed with its default parameters, and we selected the first 400 subspaces obtained by the algorithm according to (Keller et al., 2012). As for ENCLUS, we implemented the version ENCLUS_SIG, as described in (Cheng et al., 1999), since it is the faster variant of the algorithm. We also included the pruning option described by the authors to speed up the subspace analysis. The tuning of the parameters required in ENCLUS resulted in an extensive grid search over the parameter space for each dataset used in the experiments. Regarding the clustering algorithms, EWKM and AFG-*k*-means, we applied the well-known technique proposed in (Sugar and James, 2011) to set the number of clusters. In particular, for AFG-*k*-means, we used the default parameters recommended in (Gan and Ng, 2015), and the group of features per cluster delivered by the algorithm as the set of subspaces. For EWKM, we selected the attributes in each cluster with the highest weighting factor, generating as many subspaces as the number of clusters. Finally, for 4S and CMI, we followed the default parameterization suggested in the original articles.

All algorithms, with the exception of HiCS, CMI and 4S, were implemented in MATLAB® R2009b, whereas for HiCS, CMI and 4S, we made use of the publicly available code.

### 5.1.3 The Anomaly Detection Algorithm

As explained above, after executing the subspace analysis algorithm, an anomaly detection algorithm was trained on each one of the obtained subspaces. We used Minimum Volume Set (MV-Set) as presented in (Park et al., 2010) as the anomaly detection algorithm.

MV-Set, which is based on the Plug-In estimator, provides asymptotically the smallest type-II error (false negative error) for a given fixed type-I error (false positive error). More specifically, the MV-Set aims at finding the minimal support of a distribution for which the probability of each element of the support is at least as high as a predefined minimal threshold. Accordingly, the anomaly detection rule reduces to the following principle: if a new sample belongs to the minimum volume, then the new sample is considered as normal observation. Otherwise, it is labeled as abnormal. In our experiments, we used a fixed type-I error of 0.05.

Park et al. used PCA to reduce the data dimensionality before applying Kernel Density Estimation (see, e.g., Bishop, 2006) to compute the empirical probability that was later used to find the Minimum Volume set. In the experiments, Park et al. selected two principal components as it is well-known that higher dimensionality often worsens the performance of Kernel Density Estimators (Scott, 1992). We followed this approach but selected the principal components that describe 90% of the variance. If the mapped dimension of the data was found to be larger than two, then we used a Gaussian Mixture Models (GMM) (see, e.g., Bishop, 2006) to compute the supporting empirical distribution and applied it to estimate the MV-Set. The number of components in the GMM model was set to obtain a minimal Akaike Information Criterion (AIK).

We also applied another anomaly detection algorithm in our experiments, namely OC-SVM (Schölkopf et al., 1999), as a classical anomaly detection algorithm, yet, we found that in most cases, MV-Set achieved better performance, required less parameters to be tuned, and was faster to train on the same data.

## 5.2 Results

The following subsections report the detection performance results, under the three different experimental settings as described in subsection 5.1.1. Based on the experimental evaluation, we provide a detailed comparison of the proposed AAG method versus the different benchmark methods. Finally, we report the runtime that was required to train the various subspace analysis methods.

### 5.2.1 Setting 1 – Anomaly Detection (Adding Gaussian Noise)

Fig. 3 shows the resulting averaged F1-Scores as a function of the fraction of attributes synthetically perturbed by additive zero-mean Gaussian noise on six out of the 25 datasets from Table 4. In all cases, the maximum error rate $\alpha$ was set to 0.05 (see section 5.1 for more details). The x-axis indicates the fraction of perturbed attributes with respect to the total number of attributes, and the y-axis shows the averaged F1-Scores over 20 repetitions of the experiment. As seen in the figure, the proposed AAG method considerably outperforms the other methods when the fraction of perturbed attributes is lower than $\sim 0.3$. When the fraction of perturbed attributes is higher than 0.3, AAG performance remains stable, and becomes comparable to that of HiCS. Furthermore, it seems that AAG's performance is less affected by the fraction of perturbed attributes (note the lower variance in its F1-Score values), whereas the other methods are more affected by these percentages.

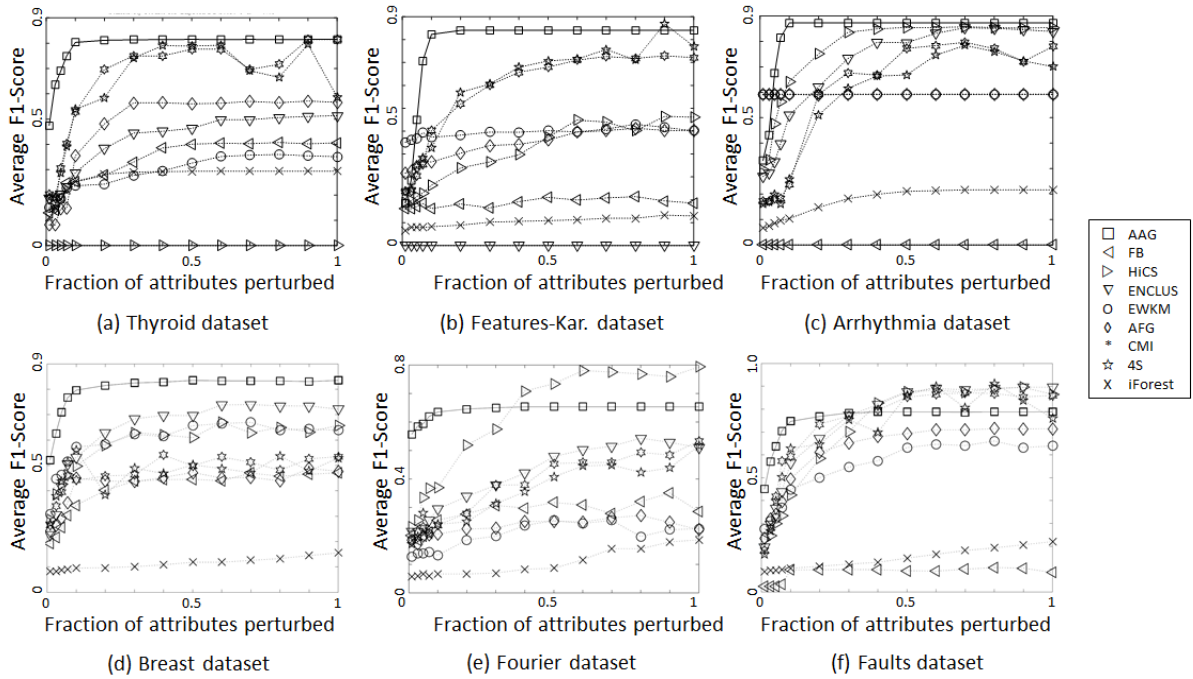

Fig. 3: Averaged F1-Score as a function of the fraction of attributes synthetically perturbed by additive zero-mean Gaussian noise, for different subspace analysis methods.

Table 6 shows the averaged F1-Scores obtained by the different subspace analysis methods, for all 25 datasets, when zero-mean Gaussian noise is added to 10% of the attributes. In each row (i.e., dataset), the two highest average F1-Score results, obtained by the two best-performing subspace analysis methods, are indicated by Bold numbers.

Table 6: Setting 1 - Averaged F1-Scores of the nine anomaly detection ensembles over the 25 UCI repository datasets. The two highest averaged F1-Scores are indicated by bold numbers.

| Dataset | AAG | FB | HiCS | ENCLUS | EWKM | AFG-$k$-means | CMI | 4S | iForest |
|---|---|---|---|---|---|---|---|---|---|
| KDDCup99(http) | 0.482 | 0.499 | 0.422 | 0.399 | 0.000 | **0.517** | 0.441 | 0.442 | **0.529** |
| KDDCup99(smpt) | **0.044** | 0.036 | 0.041 | 0.029 | 0.000 | **0.045** | 0.038 | 0.034 | 0.039 |
| Thyroid | **0.803** | 0.252 | 0.000 | 0.289 | 0.236 | 0.591 | **0.663** | 0.603 | 0.254 |
| Mammography | **0.594** | 0.579 | 0.488 | **0.598** | 0.489 | 0.212 | 0.501 | 0.473 | 0.240 |
| Glass | **0.541** | 0.376 | 0.409 | **0.553** | 0.514 | 0.375 | 0.324 | 0.324 | 0.000 |
| Breast Cancer | **0.797** | 0.344 | 0.498 | **0.532** | 0.573 | 0.449 | 0.441 | 0.445 | 0.096 |
| Zoo | 0.537 | **0.572** | 0.473 | **0.605** | 0.433 | 0.445 | 0.336 | 0.342 | 0.000 |
| Cover | 0.531 | **0.556** | 0.123 | 0.197 | 0.317 | 0.551 | **0.668** | 0.497 | 0.218 |
| Wine | **0.478** | 0.379 | 0.359 | 0.428 | 0.401 | **0.440** | 0.397 | 0.377 | 0.000 |
| Pen-Digits | **0.747** | 0.402 | 0.293 | **0.627** | 0.543 | 0.524 | 0.241 | 0.341 | 0.091 |
| Letter | 0.523 | 0.289 | 0.564 | **0.640** | 0.425 | 0.337 | 0.415 | 0.511 | 0.182 |
| Waveform 1 | 0.228 | **0.468** | **0.548** | 0.000 | 0.433 | 0.431 | 0.442 | 0.440 | 0.139 |
| Faults | **0.747** | 0.484 | 0.424 | 0.564 | 0.448 | 0.550 | **0.594** | 0.494 | 0.104 |
| Dermatology | **0.702** | 0.401 | 0.580 | **0.610** | 0.436 | 0.564 | 0.566 | 0.568 | 0.094 |
| Satimage | 0.346 | 0.186 | 0.314 | **0.365** | 0.323 | 0.303 | 0.234 | 0.239 | 0.098 |
| Waveform 2 | 0.268 | **0.637** | 0.513 | 0.573 | **0.585** | 0.513 | 0.398 | 0.387 | 0.135 |
| Segmentation | 0.720 | 0.577 | **0.733** | 0.658 | 0.608 | 0.514 | 0.605 | 0.625 | 0.000 |
| Lung Cancer | **0.753** | 0.421 | **0.704** | 0.660 | 0.448 | 0.378 | 0.627 | 0.627 | 0.000 |
| Sonar | **0.430** | 0.246 | **0.499** | 0.373 | 0.232 | 0.299 | 0.391 | 0.390 | 0.059 |
| Features Pix | 0.432 | 0.497 | 0.381 | 0.452 | **0.564** | **0.595** | 0.327 | 0.387 | 0.057 |
| Audiology | **0.712** | 0.485 | **0.674** | 0.000 | 0.370 | 0.000 | 0.492 | 0.397 | 0.000 |
| Feature Fourier | **0.635** | 0.256 | **0.370** | 0.294 | 0.147 | 0.207 | 0.238 | 0.230 | 0.067 |
| MNIST | **0.873** | **0.865** | 0.579 | 0.778 | 0.833 | 0.836 | 0.682 | 0.668 | 0.477 |
| Features Kar | **0.923** | 0.162 | 0.264 | 0.000 | 0.474 | 0.365 | **0.504** | 0.412 | 0.083 |
| Arrhythmia | **0.873** | 0.000 | **0.643** | 0.510 | 0.592 | 0.592 | 0.239 | 0.339 | 0.103 |

As seen from Table 6, in 18 out of the 25 datasets, AAG is included in the two best performing subspace analysis methods. In most of these cases, when AAG is the second best, the difference from the best method is marginal and non-significant. On the other hand, in many of the cases that AAG is ranked as the best method, the difference from the second-best method is significant. In eight cases, ENCLUS is included in the two best performing methods: in four of these cases it outperforms AAG marginally, whereas in two of these cases, AAG outperforms it significantly. In eight cases, HiCS is included in the two best performing methods: in four of these cases it outperforms AAG marginally, while AAG outperforms it in most of the cases significantly. CMI is included four times in the two best performing methods, outperforming AAG in a single dataset only ("*Cover*" dataset), and outperformed by AAG all other cases. FB is included five times among the two best performing methods, outperforming AAG in four of these cases, especially when the dataset dimensionality is relatively small. All other methods are left way behind in terms of their performance.

To further support the findings in Table 6, we conducted the statistical-significance tests that were described above. By applying the non-parametric Friedman test, we obtained an F-statistic value of $F \cong 60.747$. Based on the critical value of 3.245 at a significance level of 0.05, the null-hypothesis that all methods perform equally can be rejected.

Table 7 shows the obtained statistical values for the post-hoc tests. Based on these values, one can also reject all null-hypotheses, concluding that AAG outperforms all of its competitors in the studied cases.

Table 7: Setting 1 - Statistical significance analysis of the results from Table 6 between the proposed AAG method and the eight benchmark methods.

| Method | Rank. Diff. | z-Value | p-Value |
|---|---|---|---|
| AAG vs. FB | 2.880 | 3.718 | 0.000 |
| AAG vs. HiCS | 2.240 | 2.892 | 0.000 |
| AAG vs. ENCLUS | 1.800 | 2.324 | 0.006 |
| AAG vs. EWKM | 2.840 | 3.666 | 0.000 |
| AAG vs. AFG-*k*-means | 2.640 | 3.408 | 0.000 |
| AAG vs. CMI | 2.720 | 3.512 | 0.000 |
| AAG vs. 4S | 3.200 | 4.131 | 0.000 |
| AAG vs. iForest | 5.800 | 7.488 | 0.000 |

Fig. 4 shows the same analysis from Table 7 in form of a Critical Difference Diagram. Groups of methods that are not significantly different from each other according to the diagram are connected with a bold line. From Fig. 4 one can observe that AAG is located separately on the right side of the diagram, while all other methods (except iForest that is dominated by all other methods) are connected. This result implies again that AAG significantly outperforms all other benchmark methods.

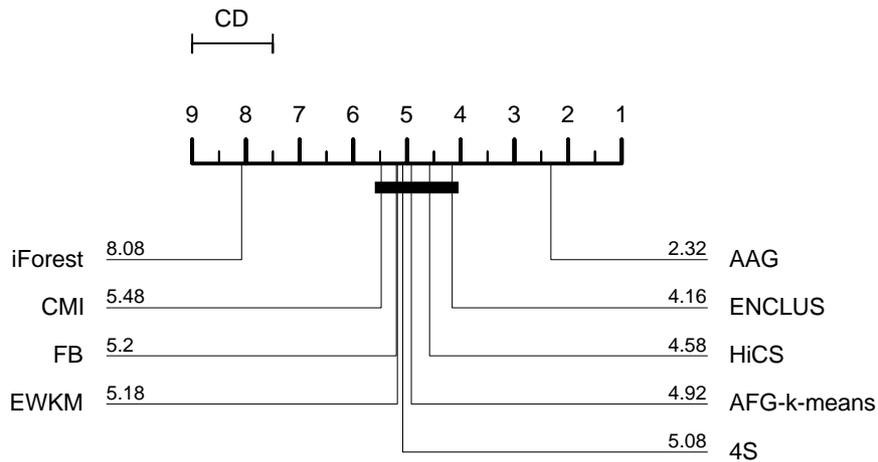

Fig. 4: Setting 1 - Comparison of all nine methods, based on the analysis from Table 7, in form of a Critical Difference Diagram. Groups of methods that are not significantly different (at a *p* value of 0.05) are connected by a bold line.

Our evaluation shows that AAG performs well at detecting anomalies when they occur in relatively small subspaces. The superiority of AAG in such cases can be explained by three main directions. First, the use of the proposed multi-attribute distance allows AAG to identify highly qualitative subspaces. Second, during the subspace combination process, AAG does not discard even a single attribute – attributes that might be necessary in the testing phase to identify anomalies that are not available for training. Third, all other benchmark methods require some tuning of parameters, where among them, one can find the number of subspaces to generate that is extremely critical. Determining the right number of subspaces is, in general, a non-trivial task, which is usually achieved by validating the framework on test data. Such a procedure may result in discarding subspaces as a result of some criterion during the training stage that can impact the performance of the anomaly detection ensemble during the testing phase, when new unseen data samples arrive.

### 5.2.2 Setting 2 – Anomaly Detection (Merging Classes)

Table 8 shows the F1-Scores obtained in the second anomaly detection setting. Recall that the reported values are averaged over 20 repetitions. Here, as well, the two best results for each dataset are indicated by Bold numbers. As seen in Table 8, in 14 out of the 25 datasets (more than any other method), AAG is included among the two best performing subspace analysis methods, and in 9 of these cases, it achieves the best performance. ENCLUS seems to be the second-best method in this anomaly detection setting – it is included among the two best performing methods in 9 of the datasets, outperforming other state-of-the-art subspace analysis methods such as HiCS, 4S and iForest. FB and AFG-*k*-means come next, both are included in the two best performing methods seven and six times, respectively. iForest is included five times in the best-two performing methods, following AAG, ENCLUS and FB. CMI is found to be less effective in detecting anomalies under this setting and is included among the two best performing methods four times only, two of them with a relatively similar performance to that of AAG. The soft subspace clustering (SSC) method EWKM is also found to be less effective than the above-mentioned methods: in only four datasets it is included among the two best performing methods. HiCS outperforms AAG in only two datasets, while 4S is not included among the two best performing methods, over all 25 datasets.

Table 8: Setting 2 - Averaged F1-Scores of the nine anomaly detection ensembles over the 25 UCI repository datasets. The two highest averaged F1-Scores are indicated by bold numbers.

| Dataset | AAG | FB | HiCS | ENCLUS | EWKM | AFG-$k$-means | CMI | 4S | iForest |
|---|---|---|---|---|---|---|---|---|---|
| KDDCup99 (http) | 0.391 | 0.299 | 0.322 | 0.331 | 0.000 | **0.427** | 0.331 | 0.332 | **0.409** |
| KDDCup99 (smtp) | **0.035** | 0.026 | 0.032 | 0.021 | 0.000 | **0.040** | 0.030 | 0.024 | 0.029 |
| Thyroid | **0.749** | 0.441 | 0.576 | 0.589 | 0.471 | 0.110 | 0.498 | 0.488 | **0.698** |
| Mammography | **0.484** | 0.479 | 0.443 | **0.498** | 0.391 | 0.200 | 0.399 | 0.381 | 0.522 |
| Glass | **0.800** | 0.345 | 0.412 | 0.426 | 0.393 | 0.314 | 0.395 | 0.212 | 0.009 |
| Breast Cancer | **0.990** | 0.986 | 0.197 | 0.217 | 0.982 | **0.985** | 0.985 | 0.903 | 0.132 |
| Zoo | 0.654 | 0.589 | **0.724** | **0.701** | 0.587 | 0.617 | 0.605 | 0.604 | 0.000 |
| Cover | 0.431 | 0.456 | 0.103 | 0.127 | 0.229 | 0.421 | **0.588** | 0.488 | **0.585** |
| Wine | 0.577 | **0.808** | 0.435 | 0.435 | **0.793** | 0.751 | 0.648 | 0.594 | 0.118 |
| Pen-Digits | 0.565 | 0.189 | 0.588 | 0.591 | 0.275 | 0.330 | 0.330 | 0.278 | **0.678** |
| Letter | 0.371 | 0.122 | 0.389 | 0.449 | 0.122 | **0.493** | 0.239 | 0.119 | 0.105 |
| Waveform 1 | 0.544 | 0.571 | 0.632 | **0.676** | 0.394 | 0.490 | 0.489 | 0.440 | 0.273 |
| Faults | **0.444** | 0.157 | 0.177 | 0.188 | **0.385** | 0.338 | 0.209 | 0.121 | 0.144 |
| Dermatology | **0.690** | 0.319 | 0.309 | 0.378 | 0.383 | **0.441** | 0.254 | 0.199 | 0.047 |
| Satimage | 0.594 | **0.596** | 0.512 | 0.522 | 0.537 | 0.556 | 0.544 | 0.546 | **0.855** |
| Waveform 2 | 0.397 | **0.617** | 0.551 | **0.565** | 0.559 | 0.401 | 0.454 | 0.455 | 0.240 |
| Segmentation | **0.741** | 0.643 | 0.715 | 0.686 | 0.549 | 0.632 | **0.708** | 0.698 | 0.000 |
| Lung Cancer | **0.835** | 0.471 | 0.658 | **0.722** | 0.638 | 0.477 | 0.535 | 0.622 | 0.296 |
| Sonar | **0.444** | **0.462** | 0.201 | 0.391 | 0.358 | 0.432 | 0.232 | 0.221 | 0.000 |
| Features Pix | 0.207 | **0.703** | 0.300 | 0.313 | **0.525** | 0.505 | 0.525 | 0.522 | 0.152 |
| Audiology | **0.776** | 0.716 | 0.702 | 0.552 | 0.503 | 0.538 | **0.826** | 0.709 | 0.071 |
| Feature Fourier | **0.370** | 0.031 | 0.166 | **0.195** | 0.025 | 0.086 | 0.175 | 0.165 | 0.134 |
| MNIST | **0.782** | **0.764** | 0.599 | 0.668 | 0.722 | 0.734 | 0.595 | 0.558 | 0.281 |
| Features Kar | 0.425 | **0.720** | 0.192 | 0.263 | 0.302 | 0.301 | 0.269 | 0.244 | 0.248 |
| Arrhythmia | 0.444 | 0.415 | 0.099 | 0.204 | **0.699** | **0.518** | 0.514 | 0.498 | 0.191 |

Again, to further support our findings in Table 8, we conducted the statistical significance tests that were described above. By applying the non-parametric Friedman test, we obtained an F-statistic value of $F \cong 32.459$. Based on the critical value of 3.245 at a significance level of 0.05, the null-hypothesis that all methods perform equally can be rejected. Table 9 shows the obtained statistical values for the post-hoc tests. Based on these values, one can also reject all null-hypotheses, concluding that AAG outperforms all of its competitors in the studied cases.

Table 9: Setting 2 - Statistical significance analysis of the results from Table 8 between the proposed AAG method and the eight benchmark methods.

| Method | Rank. Diff. | $z$-Value | $p$-Value |
|---|---|---|---|
| AAG vs. FB | 1.680 | 2.169 | 0.010 |
| AAG vs. HiCS | 2.400 | 3.098 | 0.000 |
| AAG vs. ENCLUS | 1.560 | 2.014 | 0.018 |
| AAG vs. EWKM | 2.720 | 3.512 | 0.000 |
| AAG vs. AFG-$k$-means | 1.680 | 2.169 | 0.010 |
| AAG vs. CMI | 1.720 | 2.221 | 0.008 |
| AAG vs. 4S | 3.120 | 4.028 | 0.000 |
| AAG vs. iForest | 3.840 | 4.957 | 0.000 |

Fig. 5 shows the same analysis from Table 9 in form of a Critical Difference Diagram. Groups of methods that are not significantly different from each other according to the diagram are connected with a bold line. From Fig. 5 one can observe that AAG is located separately on the right side of the diagram, while all other methods are connected. This result implies again that AAG significantly outperforms all other benchmark methods.

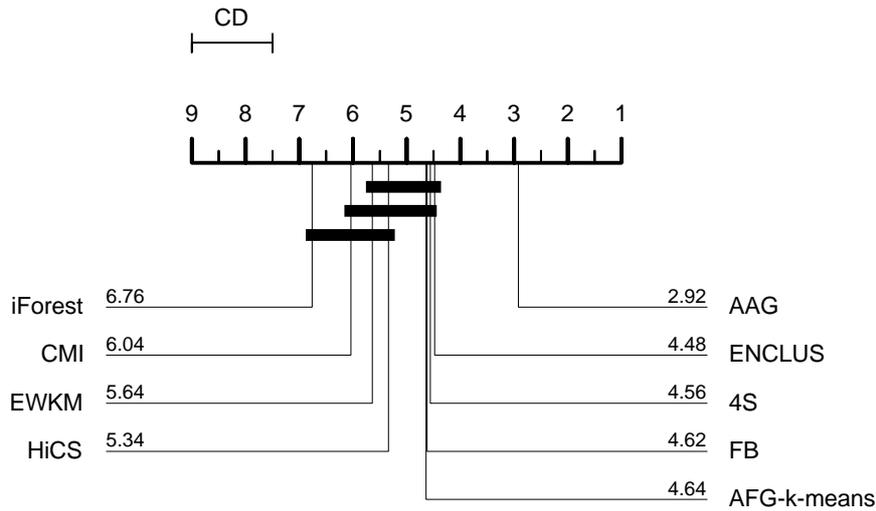

Fig. 5: Setting 2 - Comparison of all nine methods, based on the analysis from Table 9, in form of a Critical Difference Diagram. Groups of methods that are not significantly different (at a *p* value of 0.05) are connected by a bold line.

### 5.2.3 Setting 3 – Novelty Detection

Table 10 shows the averaged F1-Scores obtained in the novelty detection setting. As seen from Table 10, in 15 out of the 25 datasets, AAG is included among the two best performing subspace analysis methods, and in 6 cases, it achieves the best performance. CMI seems to be the second-best method in the novelty detection setting, being included 9 times among the two best performing methods, while in 6 of these cases it is either very close to AAG or underperforms it. AFG-*k*-means comes next, being included 6 times among the two best performing methods, with a relatively close performance of AAG in three of these cases. HiCS follows next, being included five times among the two best performing methods, with a relatively close performance of AAG in three of these cases. ENCLUS, iForest and EWKM were found to be less effective in detecting novelties and were included among the two best performing methods five, four and four times respectively.

Table 10: Setting 3 - Averaged F1-Scores of the nine anomaly detection ensembles over the 25 UCI repository datasets. The two highest averaged F1-Scores are indicated by bold numbers.

| Dataset | AAG | FB | HiCS | ENCLUS | EWKM | AFG-*k*-means | CMI | 4S | iForest |
|---|---|---|---|---|---|---|---|---|---|
| KDDCup99 (http) | **0.492** | 0.301 | 0.301 | 0.330 | 0.000 | 0.407 | 0.291 | 0.288 | **0.495** |
| KDDCup99 (smtp) | **0.041** | 0.020 | **0.044** | 0.029 | 0.000 | 0.041 | 0.024 | 0.018 | 0.038 |
| Thyroid | **0.687** | 0.339 | 0.587 | 0.357 | 0.501 | 0.201 | **0.597** | 0.537 | 0.566 |
| Mammography | **0.522** | 0.379 | 0.404 | 0.505 | 0.389 | 0.218 | 0.330 | 0.395 | **0.610** |
| Glass | 0.550 | 0.441 | 0.333 | 0.504 | 0.283 | **0.575** | 0.457 | 0.412 | 0.160 |
| Breast Cancer | 0.902 | 0.396 | 0.616 | 0.655 | 0.857 | 0.891 | **0.904** | 0.901 | 0.229 |
| Zoo | **0.581** | 0.526 | 0.460 | **0.576** | 0.527 | 0.522 | 0.161 | 0.361 | 0.167 |
| Cover | 0.514 | 0.442 | 0.091 | 0.122 | 0.290 | 0.219 | **0.598** | 0.480 | **0.588** |
| Wine | 0.570 | 0.424 | 0.400 | 0.456 | 0.561 | **0.583** | 0.523 | 0.353 | 0.192 |
| Pen-Digits | 0.827 | 0.387 | 0.637 | 0.579 | 0.743 | 0.770 | **0.875** | 0.340 | 0.249 |
| Letter | 0.173 | 0.337 | **0.553** | **0.630** | 0.275 | 0.181 | 0.169 | 0.435 | 0.407 |
| Waveform 1 | 0.634 | 0.508 | 0.602 | 0.533 | **0.746** | 0.712 | **0.728** | 0.449 | 0.299 |
| Faults | 0.377 | **0.573** | 0.448 | 0.488 | 0.394 | 0.247 | 0.236 | **0.595** | 0.291 |
| Dermatology | **0.834** | 0.578 | 0.517 | 0.460 | **0.812** | 0.770 | 0.782 | 0.619 | 0.262 |
| Satimage | **0.810** | 0.337 | 0.363 | 0.411 | **0.804** | 0.797 | 0.801 | 0.272 | 0.236 |
| Waveform 2 | 0.201 | 0.455 | 0.516 | **0.663** | 0.516 | **0.538** | 0.298 | 0.426 | 0.297 |
| Segmentation | 0.813 | 0.758 | 0.599 | 0.631 | **0.845** | **0.826** | 0.746 | 0.561 | 0.000 |
| Lung Cancer | 0.694 | 0.529 | **0.705** | 0.659 | 0.385 | 0.270 | **0.736** | 0.625 | 0.000 |
| Sonar | 0.236 | 0.305 | **0.417** | 0.349 | 0.385 | **0.453** | 0.221 | 0.393 | 0.215 |
| Features Pix | **0.855** | 0.378 | 0.432 | 0.572 | 0.531 | 0.474 | **0.792** | 0.415 | 0.233 |
| Audiology | **0.743** | 0.675 | **0.698** | 0.378 | 0.410 | 0.387 | 0.521 | 0.469 | 0.229 |
| Feature Fourier | **0.846** | 0.413 | 0.375 | 0.277 | 0.692 | 0.715 | **0.844** | 0.278 | 0.196 |
| MNIST | 0.661 | 0.677 | 0.420 | 0.441 | 0.722 | **0.735** | 0.595 | 0.600 | **0.744** |
| Features Kar | **0.846** | 0.277 | 0.204 | 0.484 | 0.577 | 0.731 | **0.794** | 0.503 | 0.262 |
| Arrhythmia | 0.468 | **0.572** | 0.495 | **0.592** | 0.495 | 0.495 | 0.431 | 0.289 | 0.240 |

Again, to further support our findings in Table 10, we conducted the statistical significance tests, as described above. By applying the non-parametric Friedman test, we obtained an F-statistic value of $F \cong 31.776$. Based on the critical value of 3.245 at a significance level of 0.05, the null-hypothesis that all methods perform equally can be rejected. **Error! Reference source not found.**Table 11 shows the obtained statistical values for the post-hoc tests. Based on these values, we can also reject all null-hypotheses, concluding that AAG outperforms all of its competitors in the studied cases.

Table 11: Setting 3 - Statistical significance analysis of the results from Table 10 between the proposed AAG method and the eight benchmark methods.

| Method | Rank. Diff. | $z$-Value | $p$-Value |
|---|---|---|---|
| AAG vs. FB | 2.5200 | 3.2533 | 0.0001 |
| AAG vs. HiCS | 2.0800 | 2.6853 | 0.0013 |
| AAG vs. ENCLUS | 1.8400 | 2.3754 | 0.0047 |
| AAG vs. EWKM | 1.6000 | 2.0656 | 0.0147 |
| AAG vs. AFG-$k$-means | 1.3200 | 1.7041 | 0.0458 |
| AAG vs. CMI | 1.5200 | 1.9623 | 0.0208 |
| AAG vs. 4S | 2.8400 | 3.6664 | 0.0000 |
| AGG vs. iForest | 3.9200 | 5.0670 | 0.0000 |

Fig. 6 shows the same analysis from Table 11 in form of a Critical Difference Diagram. Groups of methods that are not significantly different from each other according to the diagram are connected with a bold line. From Fig. 6 one can observe that AAG is located separately on the right side of the diagram, while all other methods are connected. This result implies again that AAG significantly outperforms all other benchmark methods.

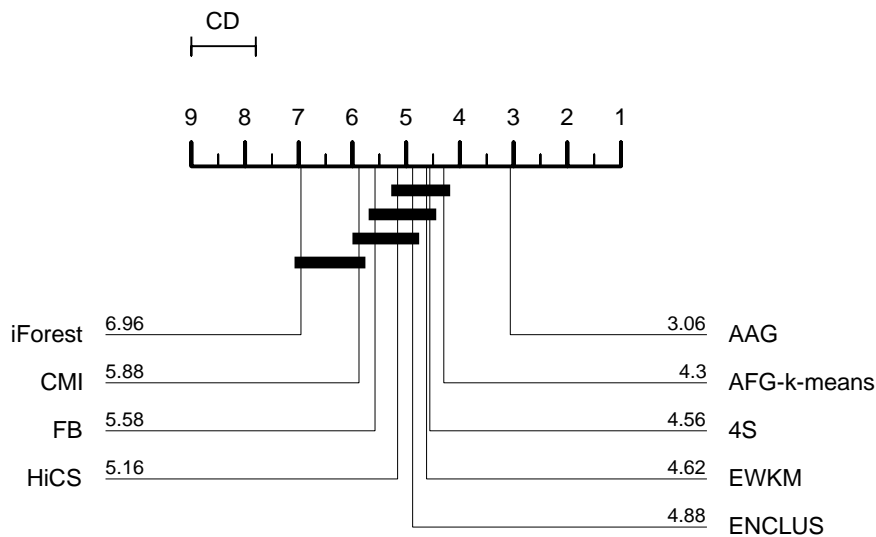

Fig. 6: Setting 3 - Comparison of all nine methods, based on the analysis from Table 11, in form of a Critical Difference Diagram. Groups of methods that are not significantly different (at a $p$ value of 0.05) are connected by a bold line.

### 5.2.4 Detailed Comparison

The rest of this subsection provides a more detailed comparison of AAG against the other benchmark methods.

With respect to the FB subspace method, the results obtained in all three settings were of relative low performance with respect to AAG. In the first two settings, FB's selection of subspaces obtained a better performance in detecting random perturbations on the attribute space as well as in detecting samples when these came from combined anomaly classes. Nevertheless, in the novelty detection setting, FB's performance was even lower. A possible reason for this might be that random combinations, as done in FB, are less prone to detect inherent correlations among different attributes that usually exhibit different data classes.

Unlike HiCS, AAG succeeds in finding a smaller number of subspaces that can be directly applied. The reason for this lies in the search strategy of HiCS, which is based on the A-Priori approach and on randomly permuted attributes to reduce the algorithm complexity. HiCS retrieves several hundreds of subspaces that afterwards have to be filtered in some way. This can be observed from the obtained results in the anomaly detection evaluation where on average, the HiCS method misses finding moderate deviations in the dataset.

With respect to ENCLUS, although it does not require to set the number of generated subspaces in advance, it does require three other parameters as input, such that their tuning requires an extensive grid search over the support of the parameters. In contrast to FB, one can see that ENCLUS often performs better in the case of anomaly detection applications (settings 1 and 2), but its performance degrades as a subspace method for novelty detection ensembles (setting 3). On the other hand, ENCLUS manages to generate a stable set of subspaces, mainly due to the A-priori search mechanism. Such a search strategy enables ENCLUS to find thousands of subspaces with a relatively small number of attributes, where several subspaces might have redundant results. Yet, due to the high number of subspaces that ENCLUS generates, potential subspaces that may find abnormal data samples are downgraded in the averaging computation of the scores. An interesting research direction might be then to evaluate different subspace combinations when the number of ensemble subspaces is high.

CMI resulted in lower performance than the proposed AAG algorithm, both for novelty detection and for anomaly detection. Nevertheless, CMI showed better results in the novelty detection setting. It seems that its subspace generation managed to combine relevant subspaces that captured the correlation among important attributes. On the other hand, 4S was not included among the two best performing subspace methods. The 4S method requires to a priori set the maximal number of attributes, which turns out to be critical for finding highly qualitative subspaces. This is manifested in the obtained results for all three examined settings.

The subspace clustering methods EWKM and AFG-k-means follow AAG, FB, HiCS and ENCLU in terms of their performance. The poorer performance with respect to all other methods is due to the fact that attributes are discarded from the set of subspaces. Consequently, neither novel nor abnormal samples can be efficiently identified. Additionally, we found that it was not trivial to set the number of clusters - a critical parameter for both methods. In both subspace-clustering methods, the number of clusters has a major impact on the selected subspaces when optimizing the extended k-means cost objective.

Finally, the iForest method achieved a poorer performance than the proposed AAG method in settings where the model is trained using only normal data and then applied to abnormal samples. Often, the iForest method is applied to outlier detection problems. That is, when abnormal and normal data samples coexist in the training dataset. It seems that only in cases where the unexpected data samples are well separated from the normal data, iForest manages to obtain a good representation of the normal data (see, e.g., Table 8 in setting 2). This may be the case when abnormal samples are almost homogeneously distributed among the subspaces that were obtained during the random training of the iForest ensemble. Nevertheless, in common real-world cases, the tree depth used to compute the threshold as anomaly score is not significant enough to generalize to unseen abnormal samples.

### 5.2.5 Runtime Evaluation

We evaluated the time taken to train each of the ensemble methods over the 25 datasets considered in this study. Since the runtimes obtained did not differ significantly among the three different settings, we show in Table 12 only the runtimes for the setting 2.

As seen in Table 12, in none of the 25 studied cases, AAG's runtime was the lowest one among the nine compared methods. HiCS and ENCLUS were found to be faster than AAG in 60% and 80% of the cases, respectively. A possible reason for this is that HiCS uses random selection of attributes to cope with runtime requirement of the original A-priori strategy. ENCLUS requires as parameter a limit to the number of attributes in each subspace and therefore, it finishes the execution even if the selected subspaces are far from optimal. As expected, FB and iForest were found to be faster than AAG in 80% and 92% of the cases, respectively, mainly, due to their random selection of attributes. Additionally, iForest does not require building an anomaly detection model over the selected subspaces, as FB does. Therefore, in most cases, iForest outperformed FB in terms of runtime.

Table 12: Averaged runtimes (in seconds) for executing the subspace analysis method and training the ensembles, for each one of the nine subspace analysis methods over the 25 studied UCI repository datasets. The two best (lowest) runtimes are indicated with Bold numbers.

| Dataset | AAG | FB | HiCS | ENCLUS | EWKM | AFG-*k*-means | CMI | 4S | iForest |
|---|---|---|---|---|---|---|---|---|---|
| KDDCup99 (http) | 131.41 | **29.31** | 304.63 | 92.25 | 111.13 | 102.02 | 142.81 | 256.52 | **56.53** |
| KDDCup99 (smtp) | 14.88 | **5.42** | 142.76 | 11.32 | 12.03 | **10.65** | 87.03 | 131.92 | 22.40 |
| Thyroid | 6.82 | 2.55 | 68.75 | **0.23** | 2.06 | **1.16** | 13.45 | 22.21 | 2.27 |
| Mammography | 25.18 | 21.77 | 43.94 | **0.11** | **1.48** | 4.25 | 22.03 | 39.38 | 4.97 |
| Glass | 1.62 | 4.16 | 1.13 | **0.35** | **0.33** | 0.47 | 0.42 | 0.71 | 0.40 |
| Breast Cancer | 0.45 | 4.02 | 3.42 | **0.07** | **0.33** | 0.50 | 0.68 | 0.90 | 0.53 |
| Zoo | 0.48 | **0.06** | 0.80 | 3.31 | 0.32 | 0.58 | 0.26 | 0.28 | **0.22** |
| Cover | 3999.70 | **0.11** | 52.86 | 103.58 | 244.67 | 402.39 | **31.43** | 36.41 | 124.21 |
| Wine | 1.67 | 3.35 | 4.60 | 2.98 | **0.56** | 0.63 | 0.77 | 0.86 | **0.27** |
| Pen-Digits | 115.22 | 23.92 | 32.98 | **0.88** | **1.72** | 3.47 | 8.26 | 11.98 | 7.92 |
| Letter | 12.44 | 4.35 | 42.47 | 3.91 | **0.82** | **1.56** | 7.13 | 7.25 | 6.81 |
| Waveform 1 | 148.51 | 8.25 | 23.99 | 30.00 | **1.74** | **2.53** | 5.24 | 5.73 | 3.62 |
| Faults | 31.83 | 2.78 | 4.58 | 36.23 | **0.51** | **1.25** | 2.99 | 4.07 | 2.52 |
| Dermatology | 7.14 | 3.43 | 3.57 | **0.46** | **0.38** | 0.55 | 1.14 | 1.68 | 0.83 |
| Satimage | 84.55 | 26.74 | 28.35 | 10.01 | **3.59** | 5.77 | 19.24 | 31.35 | **4.80** |
| Waveform 2 | 545.14 | 15.57 | 29.39 | 179.17 | **3.10** | 4.40 | 9.41 | 16.33 | **3.63** |
| Segmentation | 0.83 | 2.07 | 94.07 | 0.69 | **0.47** | 0.56 | 23.08 | 37.33 | **0.27** |
| Lung Cancer | 6.79 | 2.71 | 1.17 | 1.03 | 0.66 | **0.57** | 0.65 | 0.67 | **0.25** |
| Sonar | 37.60 | 2.49 | 11.88 | 84.55 | **0.25** | 0.41 | 2.24 | 3.41 | **0.27** |
| Features Pix | 215.06 | 7.43 | 18.98 | 18.41 | **1.28** | **1.86** | 13.13 | 20.91 | 3.19 |
| Audiology | 1.09 | 2.77 | 19.74 | **0.26** | 0.33 | 0.60 | 10.42 | 14.71 | **0.19** |
| Feature Fourier | 225.56 | 3.40 | 35.78 | 291.33 | **0.79** | **1.12** | 5.78 | 9.21 | 2.84 |
| MNIST | 704.04 | 72.77 | 135.85 | 37.21 | **12.93** | 21.94 | 23.10 | 28.06 | **5.12** |
| Features Kar | 334.53 | 5.43 | 57.29 | 100.25 | **2.85** | **3.35** | 19.49 | 31.48 | 6.25 |
| Arrhythmia | 2638.30 | 293.42 | 43.52 | 1086.12 | 2.27 | **2.05** | 8.76 | 10.62 | **0.40** |

From Table 12 it is challenging to characterize the runtime of the AAG as a function of the number of attributes or the number of samples, since each dataset has a different size and dimensionality. Recall that in Section 4.2, we analyzed the theoretical complexity of the proposed AAG method, which was found to be $O(Np^3 \log p)$, where $p$ denotes the number of attributes, and $N$ denotes the number of samples in the dataset. Figure 7 shows the relation between the theoretical complexity over each dataset and the measured runtime in practice. As expected, one can see that the empirical runtime of AAG over the 25 datasets follows the theoretical results.

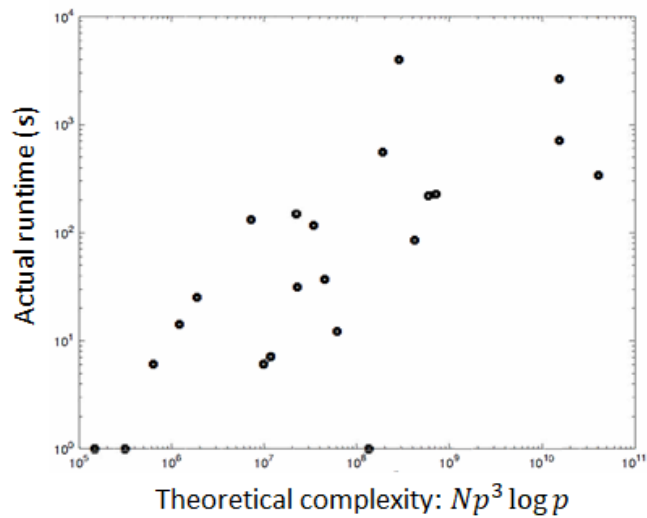

Fig. 7: Runtime evaluation of the training phase of AAG over the 25 studied datasets from the UCI repository. The y-axis represents the runtime in seconds (on a logarithmic scale), whereas the x-axis represents the theoretical complexity of the algorithm (on a logarithmic scale).

To conclude this section, Fig. 8 presents each one of the nine compared methods as a single point on a chart with two dimensions: the method's median training runtime over the 25 studied datasets, and the method's median F1-Score over the 25 datasets. As can be seen from the figure, AAG outperforms all benchmark methods in terms of F1-Score, yet this superiority comes in the expense of a longer runtime. Clearly, this is not a major issue in most real-world scenarios since the training phase is performed only once (and typically in an offline manner).

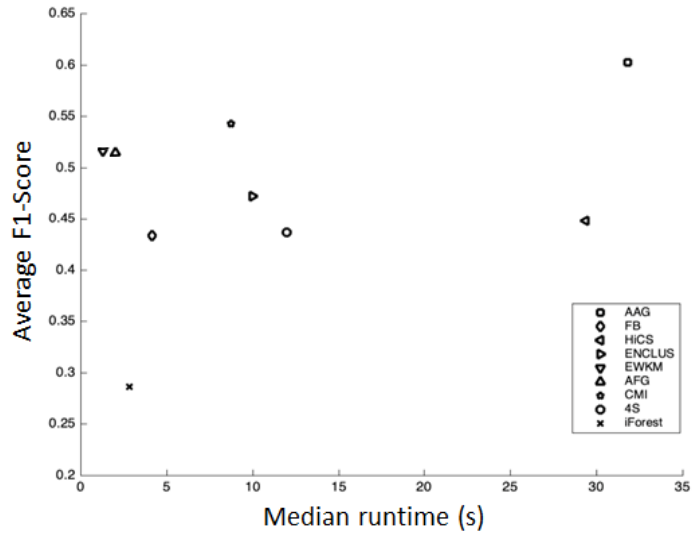

Fig. 8: Averaged F1-Score versus median runtime over all 25 studied datasets for the proposed AAG method and each of the eight other benchmark methods.

# 6 Summary and Future Work

In this paper, we introduced the Agglomerative Attribute Grouping (AAG) subspace analysis algorithm that aims at finding high-quality subspaces for anomaly detection ensembles. Similar to other state-of-the-art methods for subspace analysis, AAG searches for subspaces with highly correlated attributes. In order to assess how correlative a subset of attributes is, AAG proposes a novel measure, which was derived from previous information-theory measures over sets of partitions. We then suggest a method to approximate the proposed measure in cases where the number of attributes is large. Equipped with the newly suggested measure, AAG applies a variation of the well-known agglomerative algorithm to search for highly correlated subspaces. Our variation of the agglomerative algorithm also applies a pruning rule that reduces the potential redundancy in the final set of subspaces.

As a result of combining the agglomerative approach with the suggested measure, AAG avoids any tuning of parameters when generating the subspaces. Moreover, based on an extensive empirical study, we show that AAG outperforms other classical and state-of-the-art subspace analysis algorithms, specifically when it was used for ensemble-based anomaly detection (experimental settings 1 and 2). In both experimental settings we found that AAG training time is lower and that it can better distinguish between normal and abnormal observations. Finally, AAG also outperformed other subspace analysis methods when it was used for ensemble-based novelty detection (experimental setting 3). That is, when new classes that were not present during the training stage of the ensemble, arise in the testing stage.

While AAG demonstrated a faster training time than other state-of-the-art algorithms, its runtime complexity is proportional to $p^3$ where $p$ is the number of attributes (as analyzed in section 4.2). This property can impose a serious limitation for datasets with a very large number of attributes.

In the first anomaly detection setting, where random noise was added to normal observations), AAG obtained considerably better results than the other benchmark methods, specifically when noise was added to a relatively small number of attributes. However, when the noise was added to the entire data space, AAG lost its superiority. In cases where noise was assumed to be spread sporadically over all attributes, it might be better to use simpler anomaly detection algorithms (perhaps not even ones that are based on ensembles), to gain faster runtimes.

Recall that AAG searches for highly correlated subspaces, but it does not necessarily find the optimal set of subspaces for two main reasons: (i) the computed measure for a subset with more than two attributes is approximated; and (ii) the agglomerative algorithm is inherently a greedy one. It would be interesting to analyze the optimality boundaries obtained by AAG, and explore whether certain variations of it may result in better performance boundaries.

When preparing the datasets for the novelty detection task (experimental setting 3), we randomly sampled 10% of the minority classes that were only added to the test set. It would be interesting to experiment with other sample percentages and to analyze their impact on the detection performance of the trained ensembles.

AAG addresses the case where no separation is made between normal observations (i.e., there exists only one normal class). More specifically, in settings 1 and 3, all normal observations are taken from a single class, and in setting 2, although the normal observations can be taken from multiple classes, they are unified into a single normal class, and the separation between the underlying classes is not transparent to the algorithm. In future work, we aim to extend AAG's usage to datasets with multi-class normal observations. While the trivial way of doing so is to apply AAG on each one of the normal classes separately (and unify the sets of subspaces), we would like to utilize jointly the information available in the different classes to find higher quality subspaces.

Finally, another research direction that we plan to pursue is extending AAG to find subspaces in dynamic environments, where the probability distribution of the normal observations may change over time. Under such a scenario, we intend to first find a base set of subspaces and then to update this set incrementally when new normal observations become available.

## Acknowledgment

This research was partially supported by the MAGNET / METRO450 Consortium (http://www.metro450.org.il/) as well as by the Koret Foundation grant for Digital Living 2030.

# Appendix A: Capturing Non-Linear Relationships

To illustrate the ability of information theoretical measures to capture non-linear relationships between two random variables, we have synthetically generated data in the plane. In particular, the figure below shows three circular areas (demonstrating a non-linear relationship), that represent three different data classes. We have selected the ring in the middle and computed both the Pearson correlation and the normalized mutual information, also called, Symmetric Uncertainty, i.e., $SU(X,Y) = 2 \cdot I(X,Y) / (H(X) + H(Y))$, where $I(X,Y)$ denotes the mutual information between variables X and Y, and $H(\cdot)$ denotes the Shannon entropy.

As can be seen from the figure, in this example, Pearson correlation obtained a value of 0.11 while Symmetric Uncertainty obtained a value of 0.44. The considerably higher value obtained by the SU information theoretic measure demonstrates its ability to better capture the nonlinear relationship among these variables.

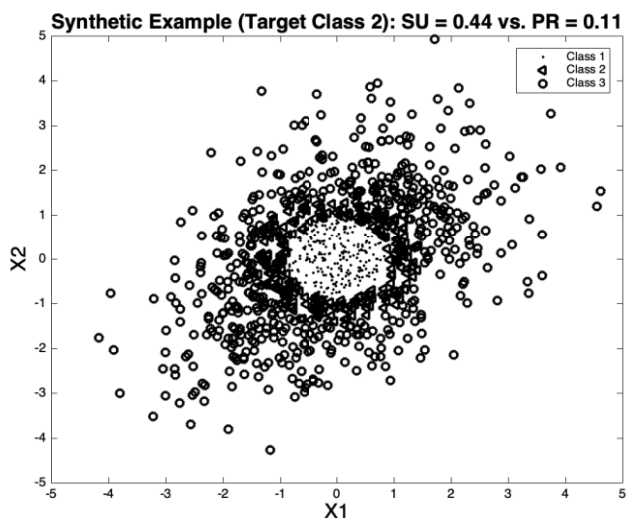

Fig. 9 Capturing non-linear relationships between two random variables.

# Appendix B: Proofs of Lemmas

## B.1 Proof of Lemma 1

**Lemma 1.** $A_j \subseteq A_i \Rightarrow d_{MA}(A_j) \geq d_{MA}(A_i)$.

**Proof:** Without loss of generality, we assume that $A_j = \{A_{j1}, A_{j2}, \ldots, A_{jk}\}$ and, $A_i = \{A_1, A_2, \ldots, A_{j1}, \ldots, A_{jk}, \ldots A_p\}$ where $A_{ji} = A_i, \forall 1 \leq i \leq k$. For simplicity, let us define $A_{\neq i} = A_i \setminus A_i$, and $A_{\neq j} = A_j \setminus A_i$, where $A_i \in A_i$, and $A_i \in A_j$. We know that $H(A_i|A_{\neq i}) \leq H(A_i|A_{\neq j})$ since conditioning the entropy cannot increase its value (Cover and Thomas, 2006). On the other side, $II(A_1, A_2, \ldots, A_p) = H(\cap_{i=1}^{p} \tilde{A}_i)$, where $\tilde{A}_i$ denotes the abstract set derived from the attribute $A_i$ and $H(\cdot)$ denotes the Shannon entropy (Reza, 1994). We simplify the notation as $H(\cap_{i=1}^{p} A_i)$ because the abstract set is constructed by the partitions generated by the attributes' values. $H(\cap_{i=1}^{p} A_i) \leq H(\cap_{j=1}^{k} A_j), \forall k \leq p$ because $\cap_{i=1}^{p} A_i$ is a decreasing sequence of $p$ sets. Therefore, $d_{MA}(A_j) \geq d_{MA}(A_i)$, for any $k \leq p$ ∎

## B.2 Proof of Lemma 2

**Lemma 2:** *Given two subspaces $A_i$ and $A_j$, such that $|A_i| \geq 2$ and $|A_j| \geq 2$, and $A_i \cap A_j = \emptyset$, then necessarily $TC(A_i \cup A_j) \geq TC(A_i) + TC(A_j)$.*

**Proof:** Assuming $|A_i| = d$ and $|A_j| = m$ and applying the definition of Total Correlation (Watanabe, 1960), we get:

$$TC(A_i \cup A_j) = TC(A_1, A_2, \ldots, A_d, A_{d+1}, \ldots, A_{d+m})$$
$$= \sum_{l=1}^{d+m} H(A_l) - H(A_1, \ldots, A_d, A_{d+1}, \ldots, A_{d+m})$$
$$= \sum_{l=1}^{d} H(A_l) + \sum_{l=d+1}^{d+m} H(A_l) - H(A_{d+1}, \ldots, X_{d+m}|A_1, \ldots, A_d) - H(A_1, \ldots, A_d)$$

since conditioning does not increase the Entropy

$$\geq \sum_{l=1}^{d} H(A_l) - H(A_1, \ldots, A_d) + \sum_{l=d+1}^{d+m} H(A_l) - H(A_{d+1}, \ldots, A_{d+m})$$
$$= TC(A_i) + TC(A_j) \qquad \blacksquare$$

# Appendix C: Stability Analysis

In this section, we analyze the stability (robustness) of the proposed AAG method as well as the other benchmark methods in yielding sets of subspaces when changes are produced in the dataset.

Often, domain experts prefer subspace analysis methods that show stability in the set of subspaces, besides acceptable performance values in detecting anomalies and novelties. Although low stability does not necessarily imply low performance rates, in many cases, low stability follows from fundamental problems in the subspace search process (e.g., Somol and Novovicova, 2010).

To proceed with the analysis, we need to first define a measure of similarity to be able to compare different methods for subspace analysis. Since the robustness analysis in selecting attributes is also related to *Feature Selection* methods in the Machine Learning community (see, e.g., Guyon et al., 2006; and Bolón-Canedo et al., 2015), we borrow definitions often applied to variable selection methods used in classification tasks. A common and useful measure of similarity used in these methods is *stability* of the attribute selection. Stability is defined as the sensitivity of a method to variations in the training dataset (Kalousis et al., 2007), and has been extensively studied with respect to the learning algorithm itself (see e.g., Křížek et al., 2006; Somol and Novovicova, 2010; Han and Yu, 2010; and García-Torres et al., 2016). Derived from the recent work presented in (García-Torres et al., 2016), we propose a way to compute the stability index of subspace analysis methods.

We denote a set of subspaces from one run of a subspace analysis method as $T_i = \{S_{i,m}\}_{m=1}^{M_i}$, where $i$ symbolizes the run-index, $M_i$ is the number of subspaces in the run $i$, and $S_{i,m}$ symbolizes one out of $M_i$ subspaces in the set $T_i$. We further denote the set of all subspaces from $L$ algorithm runs of a subspace analysis method by $\boldsymbol{S} = \{S_{i,m} \in T_i, \forall m = 1,2, \dots, M_i \text{ and } i = 1,2, \dots, L\}$.

Additionally, we denote all subspaces in $\boldsymbol{S}$ that contain $k$ attributes by $\Lambda_k$, i.e., $\Lambda_k = \{S_i, S_j \in \boldsymbol{S}: |S_i| = |S_j| = k, \forall i, j = 1,2, \dots, |\boldsymbol{S}| \text{ and } i \neq j\}$, where $|\cdot|$ denotes the cardinality of a set. Thus, $|\boldsymbol{S}|$ denotes the total number of subspaces obtained after $L$ executions of the algorithm.

Consider the following simple example. Assuming that AAG has been executed three times. Then, we obtain three sets of subspaces, i.e., $T_1$, $T_2$ and $T_3$. For simplicity, assume that, for all sets, $T_i$ for $i = 1,2,3$ comprise three subspaces, i.e., $T_i = \{S_{i,1}, S_{i,2}, S_{i,3}\}$, where $|S_{i,q}| = |S_{j,q}|$, $\forall i \neq j$ and $i,j,q = 1,2,3$. It follows that $\boldsymbol{S} = \{S_{1,1}, S_{1,2}, S_{1,3}, S_{2,1}, S_{2,2}, S_{2,3}, S_{3,1}, S_{3,2}, S_{3,3}\}$. Then, $\Lambda_2 = \{S_{1,1}, S_{2,1}, S_{3,1}\}$, $\Lambda_3 = \{S_{1,2}, S_{2,2}, S_{3,2}\}$ and $\Lambda_4 = \{S_{3,1}, S_{3,1}, S_{3,1}\}$, where we assume that $\boldsymbol{S}$ contains subspaces comprising 2, 3 and 4 attributes, respectively grouped into $\Lambda_2$, $\Lambda_3$, and $\Lambda_4$.

The approach for estimating the stability index $SI(\boldsymbol{S})$ for the set $\boldsymbol{S}$ consists of assessing the stability index for each set of equally sized subspaces, i.e., $\Lambda_k$, and then averaging the latter values. Assuming that there are $L$ sets of equally sized subspaces and each set $l = 1,2 \dots L$ is denoted as $\Lambda_{k(l)}$, where $k(l)$ refers to the number of attributes in the set $l$, then the stability index is defined as,

$$SI(\boldsymbol{S}) = \frac{1}{L}\sum_{l=1}^{L} \frac{2}{N_l(N_l - 1)} \sum_{i=1}^{N_l-1} \sum_{j=i+1}^{N_l} J(S_{i,l}; S_{j,l}) \qquad (8)$$

where $J$ is the Jaccard index, $N_l$ is the number of subspaces in the set $\Lambda_{k(l)}$, and $S_{i,l}, S_{j,l} \in \Lambda_{k(l)}$. It is easy to see that $0 \leq SI(S) \leq 1.0$, where values closer to 1.0 correspond to more stable solutions. Indeed, if all subspaces $S_{i,l}$ and $S_{j,l}$ have the same result, the double-sum term on the right side of (8) is equal to $N_l(N_l - 1)/2$. Therefore, the first sum results in $L$, computing $SI(S) = 1$.

We computed the Stability Index $SI(S)$ for the proposed AAG method, as well as for all benchmark subspaces analysis methods using the datasets described in section 5. The results shown in Table 13 are obtained for setting 1 (similar results were obtained for settings 2 and 3) after executing the corresponding subspace analysis method 20 times, where the best two results are indicated with Bold numbers.

From Table 13, we can see that, on average, the proposed AAG method, as well as the benchmark methods HiCS, ENCLUS, CMI, and 4S, achieve relatively stable solutions, whereas FB, EWKM, and AFG-*k*-means achieved less robust sets of subspaces.

A possible explanation for the lower stability of FB lies in the fact that subspaces are randomly selected. Therefore, for each algorithm run, a different set of subspaces are generated, leading to a poorer stability index. EWKM and AFG-*k*-means select subspaces by minimizing a distortion function that involves the Euclidean distance. Thus, changes in the dataset produced by the shuffling process have higher impact than their competitors, leading to a relatively lower stability in the generated subspaces.

Methods based on inherent information within the dataset suffer less from variations in the dataset. In particular, for the methods HiCS and ENCLUS, we found that the high number of selected subspaces contributes to the stability index. Specifically, both methods are based on the A-priori mechanism, and henceforth, both methods tend to select several hundred subspaces, where a small portion of attributes differs among subspaces.

Nevertheless, HiCS results are less robust than ENCLUS due to two reasons. First, it includes a random permutation of attributes to overcome the time-consuming A-priori search. Second, only the first few hundred generated subspaces are usually selected, negatively impacting the overall stability index. CMI and 4S were more robust to changes in the dataset with respect to the previously mentioned algorithms but still fall behind the proposed AAG method in stability. Recall that CMI applies the *k*-means clustering to compute the conditional mutual information, and therefore, the random dataset shuffling produces deterioration in the stability index. The 4S method, for its part, selects a specific number of attributes after computing the Total Correlation, and henceforth, the stability index shrinks. The pseudo-metric used in the search for subspaces in AAG was less influenced by the shuffling mechanism, leading to subspaces comprising almost the same attributes.

Table 13: Averaged Stability Index $SI(S)$ for each one of the nine subspace analysis methods over the 25 studied UCI repository datasets. The two best results are indicated with Bold numbers.

| Dataset | AAG | FB | HiCS | ENCLUS | EWKM | AFG-$k$-means | CMI | 4S |
|---|---|---|---|---|---|---|---|---|
| KDDCup99 (http) | **1.000** | **1.000** | **1.000** | **1.000** | **1.000** | **1.000** | **1.000** | **1.000** |
| KDDCup99 (smtp) | **1.000** | **1.000** | **1.000** | **1.000** | **1.000** | **1.000** | **1.000** | **1.000** |
| Thyroid | **0.651** | 0.369 | 0.367 | 0.537 | 0.423 | 0.466 | **0.611** | 0.601 |
| Mammography | 0.821 | 0.766 | 0.801 | 0.799 | **1.000** | **1.000** | 0.876 | 0.812 |
| Glass | 0.653 | 0.478 | **0.667** | **0.655** | 0.309 | 0.291 | 0.622 | 0.631 |
| Breast Cancer | **0.501** | 0.166 | 0.449 | 0.477 | 0.422 | 0.466 | **0.498** | 0.487 |
| Zoo | **0.713** | 0.685 | **0.691** | 0.685 | 0.408 | 0.419 | 0.644 | 0.635 |
| Cover | 0.823 | 0.732 | 0.804 | 0.813 | 0.987 | **0.998** | 0.798 | 0.809 |
| Wine | 0.617 | **0.693** | 0.583 | **0.621** | 0.289 | 0.313 | 0.587 | 0.601 |
| Pen-Digits | 0.618 | 0.580 | **0.694** | **0.622** | 0.422 | 0.458 | 0.602 | 0.611 |
| Letter | 0.549 | 0.467 | 0.488 | 0.550 | 0.340 | 0.411 | **0.610** | **0.590** |
| Waveform 1 | **0.526** | 0.423 | 0.490 | 0.443 | 0.211 | 0.190 | 0.429 | **0.511** |
| Faults | **0.595** | 0.269 | 0.475 | 0.521 | 0.201 | 0.383 | **0.588** | 0.570 |
| Dermatology | **0.652** | 0.269 | 0.521 | **0.589** | 0.267 | 0.390 | 0.499 | 0.511 |
| Satimage | **0.581** | 0.289 | 0.510 | **0.577** | 0.402 | 0.431 | 0.544 | 0.561 |
| Waveform 2 | **0.579** | 0.353 | 0.504 | **0.561** | 0.166 | 0.207 | 0.522 | 0.535 |
| Segmentation | **0.598** | 0.152 | 0.447 | 0.554 | 0.338 | 0.298 | **0.590** | 0.578 |
| Lung Cancer | **0.507** | 0.303 | 0.407 | 0.487 | **0.479** | 0.471 | 0.402 | 0.446 |
| Sonar | 0.527 | 0.297 | 0.388 | 0.601 | 0.332 | 0.378 | **0.611** | **0.612** |
| Features Pix | **0.691** | 0.106 | 0.359 | **0.609** | 0.231 | 0.233 | 0.579 | 0.591 |
| Audiology | **0.477** | 0.290 | 0.391 | **0.522** | 0.112 | 0.134 | 0.378 | 0.401 |
| Feature Fourier | **0.541** | 0.210 | 0.466 | 0.476 | 0.129 | 0.142 | 0.489 | **0.493** |
| MNIST | 0.609 | 0.123 | 0.434 | **0.655** | 0.589 | 0.609 | 0.590 | **0.612** |
| Features Kar | **0.509** | 0.151 | 0.472 | **0.510** | 0.148 | 0.201 | 0.465 | 0.490 |
| Arrhythmia | 0.573 | 0.237 | **0.583** | 0.576 | 0.281 | 0.229 | **0.579** | 0.565 |

We also performed statistical significance tests. By applying the non-parametric Friedman test, we obtained an F-statistic value of $F \cong 100.03$. Based on the critical value of 3.245 at a significance level of 0.05, the null-hypothesis that all methods performed equally can be rejected. Table 14 shows the obtained statistical values for the post-hoc tests. Differently from previous analyzed cases, we see that AAG and ENCLUS behave similarly with respect to the stability index ($p$-value > 0.05). Although both methods generate stable sets of subspaces, ENCLUS generates several hundred more subspace combinations, some of which can be redundant in the final ensemble. On the other hand, we have seen that the proposed AAG outperformed ENCLUS when it is used as subspace analysis for the ensemble of novelty detection. Finally, AAG generates, on average, relatively less subspaces than ENCLUS. This might be more attractive from the perspective of domain-expert applications, such as process monitoring.

Table 14: Statistical significance analysis of the stability results from Table 13 between the proposed AAG method and the eight benchmark methods.

| Method | Rank. Diff. | $z$-Value | $p$-Value |
|---|---|---|---|
| AAG vs. FB | 4.320 | 6.2354 | < 0.0001 |
| AAG vs. HiCS | 2.440 | 3.5218 | < 0.0000 |
| AAG vs. ENCLUS | 1.040 | 1.5011 | 0.0550 |
| AAG vs. EWKM | 4.520 | 6.5241 | < 0.0001 |
| AAG vs. AFG-$k$-means | 4.000 | 5.7735 | < 0.0001 |
| AAG vs. CMI | 2.200 | 3.1754 | < 0.0001 |
| AAG vs. 4S | 1.800 | 2.5981 | 0.0005 |